\documentclass[letterpaper]{article} 
\usepackage[preprint]{visualswap-preprint}
\usepackage[hyphens]{url}  
\usepackage{graphicx} 
\usepackage{natbib}  
\usepackage{caption} 
\usepackage{algorithm}
\usepackage{algorithmic}

\usepackage{newfloat}
\usepackage{listings}
\DeclareCaptionStyle{ruled}{labelfont=normalfont,labelsep=colon,strut=off} 
\floatstyle{ruled}
\newfloat{listing}{tb}{lst}{}
\floatname{listing}{Listing}

\usepackage{booktabs}
\usepackage{multirow}
\usepackage{array}
\usepackage{amssymb}

\title{Recompute or Reuse? Diagnosing and Mitigating Textual Shortcuts in VLM Self-Reflection}
\author {
    Wenxiao Fan\textsuperscript{\rm 1}\thanks{This work was conducted while Wenxiao Fan was an intern at JD.com.},
    Jingling Fu\textsuperscript{\rm 2}, 
    Fang Li\textsuperscript{\rm 2}, 
    Luohang Liu\textsuperscript{\rm 2}, 
    Yu He\textsuperscript{\rm 2}, 
    Lichen Ma\textsuperscript{\rm 2,3}, 
    Zhiyang Yu\textsuperscript{\rm 2}, 
    Weishan Bi\textsuperscript{\rm 2}, 
    Junshi Huang\textsuperscript{\rm 2}, 
    Yan Li\textsuperscript{\rm 2}, 
    Gu Simiu\textsuperscript{\rm 2}, 
    Kan Li\textsuperscript{\rm 1}\corresponding
}
\affiliations {
    \textsuperscript{\rm 1}School of Computer Science, Beijing Institute of Technology\\
    \textsuperscript{\rm 2}JD.COM\\
    \textsuperscript{\rm 3}Institute of Artificial Intelligence and Robotics, Xi'an Jiaotong University\\
    \{wenxiaofan, likan\}@bit.edu.cn
}

\begin{document}

\maketitle

\begin{abstract}
Vision-language models (VLMs) are expected to revise their reasoning when visual evidence changes. Failures to do so are often attributed to insufficient visual attention or contextual inertia, leaving unclear what models reuse instead of recomputing from the current image. We show that evidence-bearing reasoning in a prior chain of thought (CoT) can form a textual shortcut that competes behaviorally with visual recomputation. Across 16 VLMs, a matched counterfactual analysis identifies evidence-bearing content as the most robust carrier of prior-CoT influence. 
Removing this evidence-bearing content shifts answer preference more than removing length-matched non-evidence context or the final-answer span, with prior control weakening progressively as more stale evidence is removed.
Reordering this evidence also weakens prior control, showing that its organization modulates shortcut strength.
Beyond the immediate answer, the shortcut can retain residual influence after answer correction: weakening current-image support shifts preference back toward the prior answer, while repeated prior answers and reused premises arise mainly when the shortcut remains active. To limit this influence, we introduce Fresh-State Attention Firewall (FSAF), a training-free intervention that isolates fresh computation from the prior CoT. Across five VLMs, FSAF raises visual update rate from \(35.28\%\) to \(53.61\%\) and reduces prior-answer rate from \(39.22\%\) to \(3.67\%\). Reliable VLM self-reflection therefore requires more than looking again: fresh visual recomputation must be protected from stale textual reuse.
\end{abstract}

\section{Introduction}

Vision-language models (VLMs) are increasingly expected to inspect their
reasoning and revise answers when visual evidence changes, extending the
broader paradigm of self-feedback and self-correction
\citep{madaan2023selfrefine,kamoi2024when}. Reliable self-reflection
requires invalidating evidence from the previous visual state and
recomputing from the current image, not merely replacing one answer with
another. VisualSwap shows that VLMs often fail to do so even when they
claim to re-examine the image \citep{shi2026visualswap}. Accordingly,
recent methods encourage renewed visual attention
\citep{jian2025lookagain,yang2026lookback}. This
diagnosis explains what is missing, but not what takes its place in the
computation. We therefore ask: \textit{when a VLM fails to recompute,
what does it reuse instead?}

\begin{figure}[t]
    \centering
    \includegraphics[width=\columnwidth]{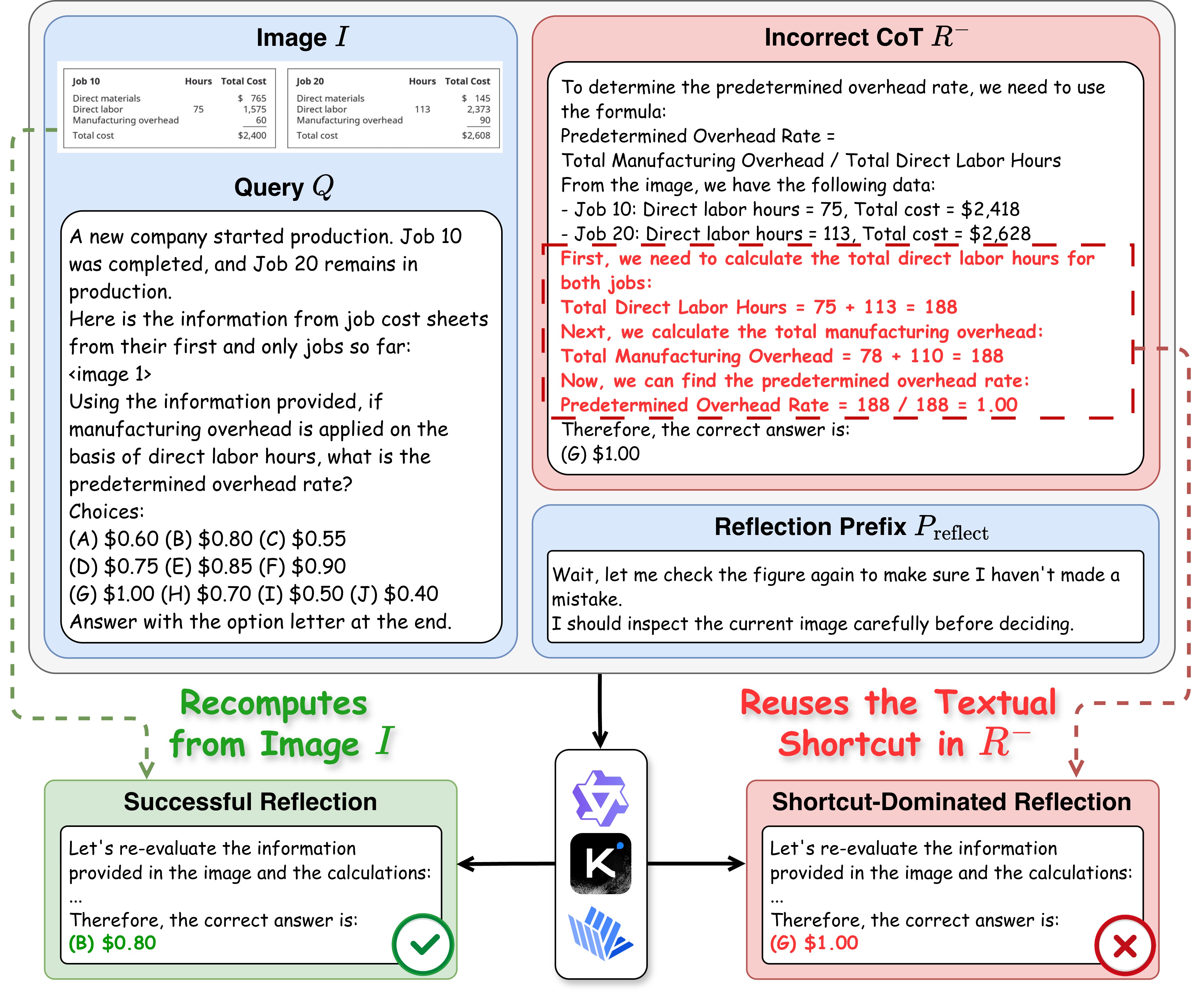}
    \caption{\textbf{Textual shortcut during VLM self-reflection.}
    Reusing this prior evidence path forms a textual
    shortcut that can steer the reflected answer without fresh
    recomputation from the current image.
    }
    \label{fig:intro-recompute-reuse}
    \vspace{-1.2em}
\end{figure}

We argue that prior reasoning can become an alternative computation
path. A prior CoT does not merely record a final answer; it textualizes
observations, numerical values, relations, and intermediate deductions
grounded in the previous image. After the image changes, this
evidence-bearing path remains available in context. As illustrated in
Fig.~\ref{fig:intro-recompute-reuse}, the model can either reconstruct
the relevant evidence from the current image or reuse a path already
organized in text. We call the latter a
\emph{textual shortcut}. This reframes visual self-reflection as a
competition between fresh visual recomputation and stale textual reuse.

\begin{figure*}[t]
    \centering
    \includegraphics[width=\linewidth]{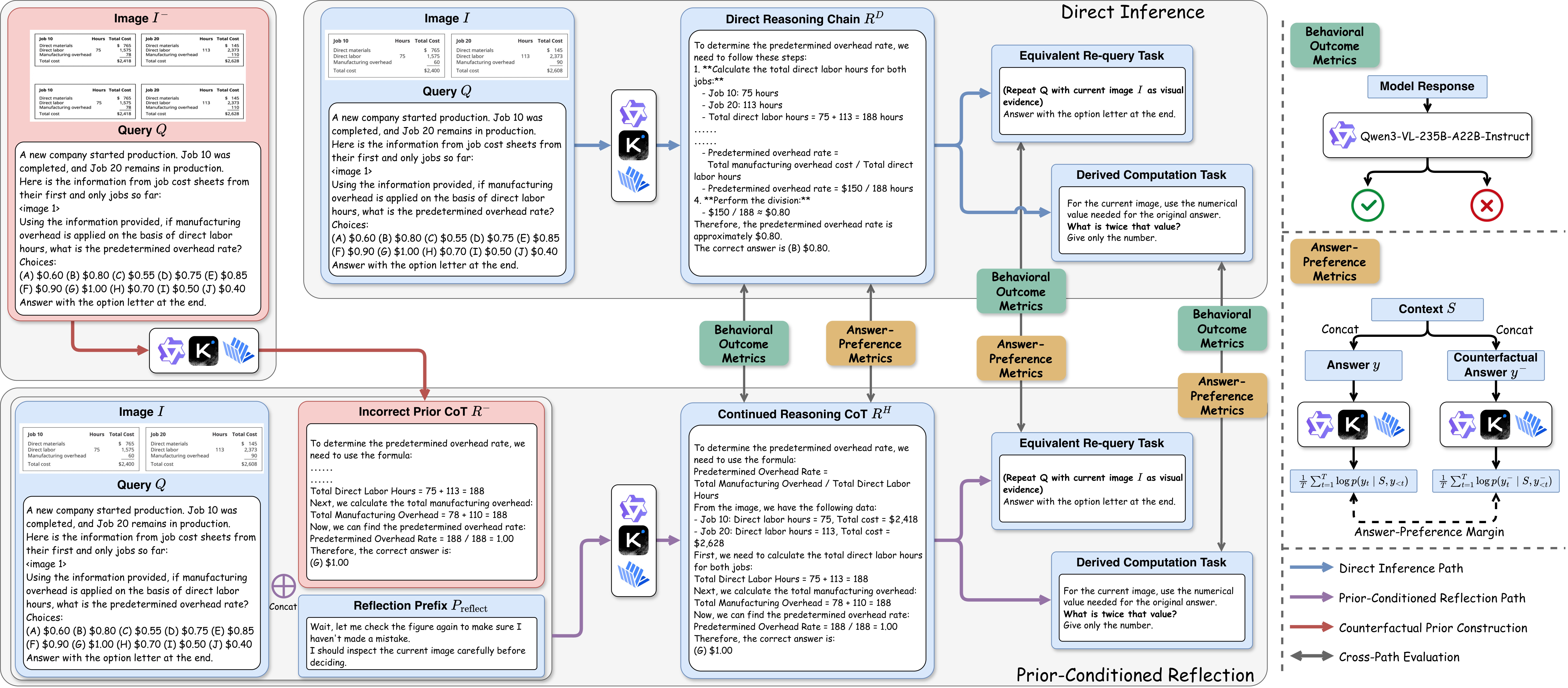}
    \caption{\textbf{Controlled evaluation framework.}
    A matched counterfactual image elicits a coherent, model-authored
    incorrect prior CoT (red). Under the current image, we compare direct
    inference (blue) with prior-conditioned reflection (purple), including
    their initial responses and evidence-preserving continuations (gray),
    using behavioral and answer-preference metrics.}
    \label{fig:framework}
    \vspace{-0.2em}
\end{figure*}

We study this competition with a matched counterfactual framework
(Fig.~\ref{fig:framework}). Each question is paired with two structurally
similar images that imply different answers. The model first produces a
coherent, model-authored incorrect prior CoT \(R^{-}\) from the
counterfactual image \(I^{-}\), then reflects under the current image
\(I\). Comparing direct and prior-conditioned paths measures the prior
CoT's influence, while controlled removals separate evidence-specific
control from context reduction and final-answer anchoring.

Experiments across 16 VLMs reveal two phenomena. First,
\textbf{evidence-bearing reasoning is the most robust cross-model carrier of prior control}. 
Removing it shifts answer preference more than
length-matched non-evidence removal or final-answer removal; the
content-specific effect is positive in all 16 models and grows as more
stale evidence is removed. Disrupting evidence order also weakens prior
control, showing that organization modulates how readily the shortcut is
reused during subsequent visual self-reflection.

Second, \textbf{producing the current answer does not imply that the shortcut has retired}. Under matched support withdrawal, prior-conditioned
recovered cases shift back toward the prior answer more strongly than
history-free controls. Strong answer recurrence and premise reuse arise
mainly when the stale shortcut remains active, whereas recovered cases
primarily retain a weaker score-level pull. Answer correction is
therefore an insufficient certificate of evidence-state revision while
prior reasoning remains accessible.

This diagnosis motivates \emph{Fresh-State Attention Firewall} (FSAF), a
training-free intervention that prevents fresh computation from
attending to the prior CoT. FSAF requires neither target-answer access
nor an evidence parser. Across five Qwen VLMs, it raises visual update
rate from \(35.28\%\) to \(53.61\%\) and reduces prior-answer rate from
\(39.22\%\) to \(3.67\%\). A prompt-only variant provides no reliable
update gain, indicating that controlling access to stale reasoning,
rather than merely asking the model to look again, drives the
improvement observed across all  models.

Our contributions are threefold:
\begin{itemize}
    \item We identify evidence-bearing prior reasoning as a reusable
    textual shortcut that competes with current visual recomputation.
    Across 16 VLMs, its control is content-specific, graded with retained
    stale evidence, and modulated by evidence organization.
    \item We show that answer correction does not guarantee shortcut
    retirement: recovered cases can retain residual prior dependence,
    while strong recurrence and premise reuse are concentrated where the
    stale shortcut remains active.
    \item We introduce FSAF, a training-free intervention that isolates
    fresh computation from the prior CoT and improves visual updating
    across five Qwen VLMs.
\end{itemize}

\section{Related Work}

\noindent\textbf{Visual Re-examination and Grounded Reflection.}
Controlled visual tests and perturbations reveal that a plausible answer or
reflective statement does not guarantee sensitivity to the current image
\citep{shi2026visualswap,sun2026perceive,guan2024hallusionbench}. Longer multimodal reasoning can
further weaken visual grounding \citep{xu2025morethinking}. Accordingly,
recent methods encourage visual re-attention through reflection training,
attention-aware rewards, or explicit visual-token revisiting
\citep{jian2025lookagain,chu2025qwenlookagain,yang2026lookback,
cheng2025r3v,ji2025reflair,tang2026groundedreflection}. Training-free
alternatives counter language-prior dominance through visual contrastive
decoding or targeted hidden-state intervention
\citep{leng2024vcd,ji2026causallens}. Complementary work evaluates
whether intermediate perception steps, rather than only final answers,
remain visually faithful \citep{uppaal2026journey,zhang2026seeit}. These studies focus
on restoring current-image use. VisualSwap studies a dynamic conflict
after an image swap, whereas VI-Probe isolates a static
perception--memory conflict; both motivate evaluating responsiveness to
changed evidence rather than accuracy alone. We instead identify the
prior textual state that competes with fresh visual recomputation.

\noindent\textbf{Self-Correction and Belief Revision.}
Self-correction methods refine outputs through self-feedback, verbal
reflection, or tool-grounded critique
\citep{madaan2023selfrefine,shinn2023reflexion,gou2024critic,
pan2024automatic}. However,
intrinsic correction without reliable external feedback is often
unstable and can degrade reasoning
\citep{huang2024cannot,kamoi2024when,zhang2024darkside,
he2025selfcorrection}. Belief-revision studies further
distinguish updating when premises change from maintaining a valid
conclusion, while correction analyses expose a similar trade-off between
critique and answer retention
\citep{wilie2024beliefrevision,zhang2024selfcontrast,
yang2025confidence}. Together, these results suggest that successful
revision depends on both detecting a conflict and replacing the premises
that support the earlier conclusion. Our setting makes the changed
evidence visual and asks whether answer correction actually retires the
prior evidence path.

\noindent\textbf{Prior Reasoning as Persistent Context.}
CoT is not necessarily a faithful explanation of a model's
decision, although interventions on the trace show that it can still
causally influence the answer
\citep{turpin2023unfaithful,lanham2023faithfulness}. Conversational
history can also induce persistent generation tendencies
\citep{simhi2026oldhabits}, while correction and trust vary with whether
the same claim appears as the model's own thought, a user message, or
another role \citep{chen2026selfcorrection,pan2026userassistant}. Such
role effects characterize source preference, but do not determine which
parts of a prior reasoning trace carry task-specific control. We study a
more specific multimodal form of persistence: evidence-bearing reasoning
generated from an earlier image remains accessible as a textual shortcut
after the visual evidence changes.

\section{Controlled Evaluation of Prior-CoT Influence}
\label{sec:framework}

\noindent\textbf{Overview.}
Prior work \citep{shi2026visualswap,jian2025lookagain} shows that VLM
self-reflection can fail to re-engage current visual evidence, but
leaves open what controls the answer when fresh recomputation fails. We
introduce a matched counterfactual framework to trace prior-CoT
influence during and after self-reflection. As shown in
Fig.~\ref{fig:framework}, we hold the current task fixed and compare
history-free direct inference with
prior-conditioned reflection, both at the initial response and under
matched continuations. This tests whether current visual evidence
overrides a coherent but stale reasoning path.

\noindent\textbf{Problem Formulation.}
We pair a current task \(\mathcal{T}=(I,Q,y)\) with a matched
counterfactual \(\mathcal{T}^{-}=(I^{-},Q,y^{-})\) that shares the
question and answer space but changes the task-relevant visual evidence,
so \(y^{-}\neq y\). We instantiate this conflict with 800 human-checked
VS-Bench pairs
\citep{shi2026visualswap}, drawn equally from MathVista-MINI,
MathVerse-MINI, MathVision, and MMMU-Pro
\citep{lu2024mathvista,zhang2024mathverse,wang2024mathvision,
yue2025mmmupro}. We run the evaluated model \(\mathcal{M}\) on
\(I^{-}\):
\begin{equation}
    (R^{-},\hat y^{-})=\mathcal{M}(I^{-},Q)
    \label{eq:prior-state}
\end{equation}
Eligibility requires \(\hat y^{-}=y^{-}\) and direct inference on \(I\)
to return \(y\). Thus, \(R^{-}\) coherently supports \(y^{-}\) under
\(I^{-}\) but conflicts with \(y\) under \(I\), while both tasks remain solvable.

\noindent\textbf{Evaluation Protocol.}
For each eligible pair, the history-free \emph{direct inference path}
excludes the counterfactual prior:
\begin{equation}
    (R^{D},\hat y^{D})=\mathcal{M}(I,Q)
    \label{eq:direct-state}
\end{equation}
The \emph{prior-conditioned reflection path} receives \(I\), \(Q\),
\(R^{-}\), and a fixed reflection prefix \(P_{\mathrm{reflect}}\):
\begin{equation}
    (R^{H},\hat y^{H})
    =
    \mathcal{M}(I,Q,R^{-},P_{\mathrm{reflect}})
    \label{eq:history-state}
\end{equation}
Holding \(I\) and \(Q\) fixed, the contrast measures influence
associated with prior-conditioned history, including the fixed prefix,
rather than base task difficulty.

\noindent\textbf{Evidence-Preserving Continuations.}
We also apply this contrast to two matched
continuations (Fig.~\ref{fig:framework}). The \emph{equivalent
re-query task} repeats \(Q\) to test whether the original answer path is
reinstated; the \emph{derived computation task} applies a new operation
to test whether a preceding premise transfers into a new computation.
Both retain the current image and accumulated evidence, separating
prior-associated persistence from generic follow-up instability.

\noindent\textbf{Behavioral Outcome Metrics.}
We assess free-generation behavior and teacher-forced answer preference.
For free generation,
Qwen3-VL-235B-A22B-Instruct independently judges whether each response
\(\mathcal{M}(S_i)\) matches the current target \(y_i\) or
counterfactual target \(y_i^{-}\). Let \(J(r,a)\in\{0,1\}\) indicate a
semantic match. Over \(N\) eligible pairs, we compute
\begin{equation}
    \begin{array}{ll}
        \mathrm{VUR}(S)
        &=
        \displaystyle\frac{1}{N}
        \sum_{i=1}^{N}
        J\!\left(\mathcal{M}(S_i),y_i\right)
        \\[3pt]
        \mathrm{PAR}(S)
        &=
        \displaystyle\frac{1}{N}
        \sum_{i=1}^{N}
        J\!\left(\mathcal{M}(S_i),y_i^{-}\right)
    \end{array}
    \label{eq:vur-par}
\end{equation}
Visual update rate (VUR) measures agreement with the current answer,
whereas prior-answer rate (PAR) measures agreement with the
counterfactual answer. We distinguish three interpretable outcomes:
\emph{recovered} (current only), \emph{active-stale} (prior only), and
\emph{Other} (neither target). 
 These labels describe observed outputs, not hidden states. The same judge
establishes eligibility. Rates are reported in \% and differences in
percentage points (pp).

\begin{table}[t]
\centering
\small
\begin{tabular}{lcc}
\toprule
\textbf{Model}
& \shortstack{\(\mathbf{Q+R^{-}}\), \textbf{no} \(\mathbf{I}\)}
& \(\mathbf{I+Q+R^{-}}\) \\
& \textbf{VUR / PAR} & \textbf{VUR / PAR} \\
\midrule
Qwen3-VL-8B (Thinking) & 0.6 / 53.3 & 22.6 / 7.0 \\
Kimi-VL-A3B (Instruct)  & 1.1 / 97.2 & 56.9 / 39.8 \\
Gemma-4-31B (IT)        & 1.1 / 92.7 & 88.0 / 4.4 \\
\bottomrule
\end{tabular}
\caption{\textbf{Prior CoT can control reflection without the current image.}
Results for all 16 VLMs appear in
Tab.~\ref{tab:prior-influence-full}. 
}
\label{tab:prior-influence-main}
\vspace{-1em}
\end{table}

\begin{table*}[t]
\centering
\small
\begin{tabular*}{0.75\textwidth}{@{}llccccc@{}}
\toprule
& & \multicolumn{3}{c}{\textbf{Answer Preference}}
& \multicolumn{2}{c}{\textbf{Free-Generation Outcomes (pp)}} \\
\cmidrule(lr){3-5}\cmidrule(lr){6-7}
\textbf{Model}
& \textbf{Variant}
& \(\Delta m_{\mathrm{E-N}}\)
& \(\Delta m_{\mathrm{E-C}}\)
& \(\Delta m_{\mathrm{A-N}}\)
& \(\Delta\mathrm{VUR}_{\mathrm{E-C}}\)
& \(\Delta\mathrm{PAR}_{\mathrm{E-C}}\) \\
\midrule
Qwen2.5-VL-7B
  & Instruct & +2.019 & +1.948 & +0.064 & +46.12 & -67.05 \\
\midrule
\multirow{2}{*}{Qwen3-VL-8B}
  & Instruct & +2.699 & +2.617 & +0.237 & +42.27 & -72.95 \\
  & Thinking & +2.826 & +2.832 & +0.237 & +35.39 & -84.65 \\
\midrule
\multirow{2}{*}{Qwen3-VL-32B}
  & Instruct & +3.142 & +3.075 & +0.045 & +54.74 & -81.52 \\
  & Thinking & +2.995 & +3.004 & +0.159 & +54.47 & -82.11 \\
\midrule
Qwen3.5-4B
  & -- & +1.548 & +1.612 & +0.057 & +45.49 & -70.82 \\
\midrule
Qwen3.5-9B
  & -- & +1.345 & +1.372 & +0.033 & +47.11 & -65.29 \\
\midrule
Qwen3.5-27B
  & -- & +1.600 & +1.535 & +0.050 & +58.80 & -72.66 \\
\midrule
Qwen3.6-27B
  & -- & +1.635 & +1.591 & +0.053 & +53.53 & -71.00 \\
\midrule
InternVL2-8B
  & Instruct & +0.926 & +0.913 & +0.008 & +26.95 & -33.33 \\
\midrule
\multirow{2}{*}{Gemma-4-12B}
  & Base & +0.902 & +0.911 & +0.115 & +50.00 & -65.00 \\
  & IT   & +2.221 & +2.067 & +0.025 & +26.64 & -28.38 \\
\midrule
\multirow{2}{*}{Gemma-4-31B}
  & Base & +1.017 & +0.996 & +0.061 & +49.26 & -66.18 \\
  & IT   & +3.334 & +3.152 & -0.005 & +25.45 & -25.09 \\
\midrule
\multirow{2}{*}{Kimi-VL-A3B}
  & Instruct & +1.742 & +1.619 & -0.029 & +32.04 & -49.17 \\
  & Thinking & +3.145 & +2.913 & +0.078 & +51.91 & -66.12 \\
\midrule
\multicolumn{2}{l}{\textbf{Average}}
  & \textbf{+2.069} & \textbf{+2.010} & \textbf{+0.074}
  & \textbf{+43.76} & \textbf{-62.58} \\
\bottomrule
\end{tabular*}
\caption{\textbf{Evidence-bearing content carries textual-shortcut control.}
N is the intact prior; E removes \(E^{-}\); C is the length-matched
non-evidence context control; and A removes \(A^{-}\). Margin columns
report the indicated contrasts, while the two rightmost columns report E--C
changes in VUR/PAR. Positive margin/VUR and negative PAR
shifts favor the current answer.
}
\label{tab:evidence-content}
\vspace{-1.1em}
\end{table*}

\noindent\textbf{Answer-Preference Metrics.}
Free-generation outcomes can hide residual preference. We therefore
score each candidate \(a=(a_1,\ldots,a_{|a|})\) by its mean token
log-probability under teacher forcing:
\begin{equation}
    \ell(a\mid S)
    =
    \frac{1}{|a|}
    \sum_{t=1}^{|a|}
    \log p_{\mathcal{M}}(a_t\mid S,a_{<t}).
    \label{eq:answer-score}
\end{equation}
Following likelihood-based evaluations
\citep{malladi2023finetuning,ren2023selfevaluation,li2024multiplechoice},
we define
\begin{equation}
    m(S)
    =
    \ell(y\mid S)-\ell(y^{-}\mid S).
    \label{eq:new-old-margin}
\end{equation}
Positive (negative) \(m(S)\) favors the current (prior) answer.
This evaluated-model score is independent of the semantic judge and
assumes no unique hidden representation.

\section{Evidence-Bearing Prior CoTs Form Shortcuts}
\label{sec:textual-shortcut}

Having established the controlled prior-conditioned path, we now ask
whether \(R^{-}\) behaves as a reusable alternative to recomputing from
the current image \(I\). Such a path should exert answer control on its
own, depend specifically on evidence-bearing content rather than
generic context or the final-answer span, and weaken when its
evidential organization is disrupted.

\begin{figure}[t]
\centering
\includegraphics[width=1\linewidth]{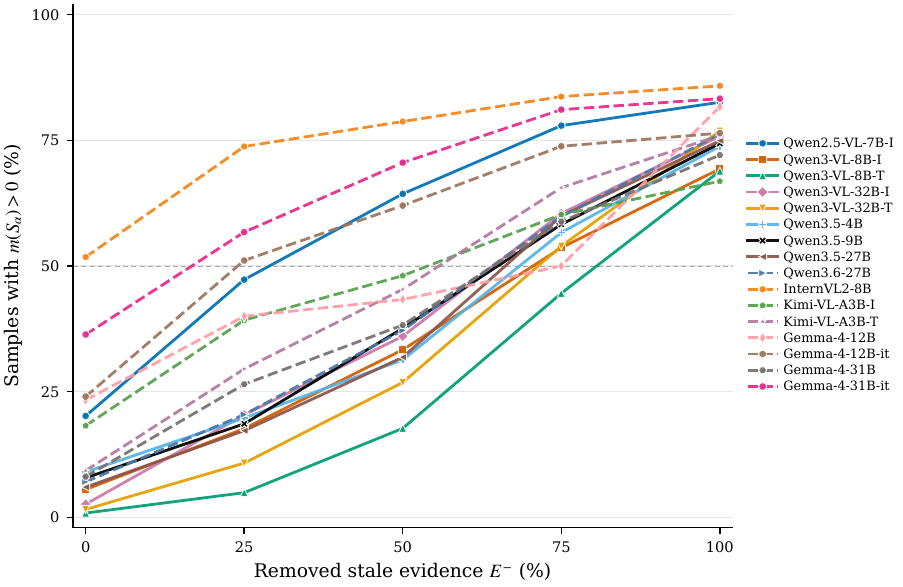}
\caption{\textbf{Evidence removal progressively weakens prior-answer control.}
The proportion of samples with
\(m(S_{\alpha})>0\) increases monotonically at the model level in all
16 VLMs. 
}
\label{fig:evidence-dose-response}
\vspace{-1em}
\end{figure}

\noindent\textbf{Prior CoT Is an Active Source of Answer Control.}
Before locating its carrier, we test whether \(R^{-}\) can itself drive
task-specific answers. We compare a \emph{prior-only}
condition (\(Q+R^{-}\), no \(I\)) with the matched
\emph{image-present} condition. Tab.~\ref{tab:prior-influence-main}
reports three representative model families. Without \(I\), VUR remains
\(0.6\%\)--\(3.5\%\), whereas PAR reaches \(53.3\%\)--\(97.2\%\),
showing that \(R^{-}\) can sustain prior-answer control. Restoring \(I\)
shifts predictions toward the current answer, but recovery remains
heterogeneous (VUR \(22.6\%\)--\(88.0\%\), PAR
\(4.4\%\)--\(53.9\%\)). Thus, \(R^{-}\) supplies a competing answer
path, but not which content carries its control.

\begin{table}[t]
\centering
\footnotesize
\setlength{\tabcolsep}{0pt}
\newcommand{\orgppfield}[1]{\makebox[3.05em][l]{#1}}
\newcommand{\orgvurpar}[2]{%
  \orgppfield{#1}\,/\,\hspace{0.08em}\orgppfield{#2}}
\begin{tabular*}{\columnwidth}{@{\extracolsep{\fill}}lccc@{}}
\toprule
\multirow{3}{*}{\textbf{Model (Variant)}} & \multicolumn{2}{c}{\textbf{Answer Preference}}
& \textbf{Free Generation (pp)} \\
\cmidrule(lr){2-3}\cmidrule(lr){4-4}
& \(\Delta m_{\mathrm{D-N}}\)
& \(\Delta m_{\mathrm{D-NE}}\)
& \(\Delta\mathrm{VUR}\,/\,\Delta\mathrm{PAR}\) \\
\midrule
Qwen2.5-VL-7B (I) & +0.645 & +0.641 & \orgvurpar{+13.15}{-15.94} \\
\midrule
Qwen3-VL-8B (I)   & +0.930 & +0.741 & \orgvurpar{+4.10}{-6.41} \\
Qwen3-VL-8B (T)   & +0.771 & +0.444 & \orgvurpar{+1.92}{-11.94} \\
\midrule
Qwen3-VL-32B (I)  & +1.141 & +0.808 & \orgvurpar{+13.27}{-15.88} \\
Qwen3-VL-32B (T)  & +0.478 & +0.446 & \orgvurpar{+3.68}{-5.79} \\
\midrule
Qwen3.5-4B        & +0.104 & +0.099 & \orgvurpar{+2.59}{-3.45} \\
Qwen3.5-9B        & +0.065 & +0.076 & \orgvurpar{+3.72}{-3.72} \\
Qwen3.5-27B       & +0.045 & +0.044 & \orgvurpar{+3.01}{-3.01} \\
Qwen3.6-27B       & +0.045 & +0.032 & \orgvurpar{+1.87}{-3.75} \\
\midrule
InternVL2-8B (I)  & +0.182 & +0.199 & \orgvurpar{+4.08}{-22.45} \\
\midrule
Gemma-4-12B (B)   & -0.049 & +0.013 & \orgvurpar{-7.50}{+10.00} \\
Gemma-4-12B (IT)  & +0.878 & +0.686 & \orgvurpar{+17.11}{-9.65} \\
Gemma-4-31B (B)   & +0.138 & -0.016 & \orgvurpar{+6.78}{-7.63} \\
Gemma-4-31B (IT)  & +1.378 & +0.673 & \orgvurpar{+17.88}{-12.41} \\
\midrule
Kimi-VL-A3B (I)   & -0.138 & -0.155 & \orgvurpar{+0.00}{+0.00} \\
Kimi-VL-A3B (T)   & +0.565 & +0.343 & \orgvurpar{+7.73}{-8.84} \\
\midrule
\textbf{Average}
  & \textbf{+0.449} & \textbf{+0.317}
  & \textbf{\orgvurpar{+5.84}{-7.55}} \\
\bottomrule
\end{tabular*}
\caption{\textbf{Disrupting evidence-chain organization weakens
textual-shortcut control.}
D/N/NE denote disrupted, intact, and non-evidence-reordered priors;
I/T/B/IT denote Instruct/Thinking/Base/instruction-tuned. The last
column reports D--N \(\Delta\mathrm{VUR}/\Delta\mathrm{PAR}\) (pp).
}
\label{tab:evidence-organization}
\vspace{-1.2em}
\end{table}

\noindent\textbf{Evidence-Bearing Content Carries Textual Shortcut Control.}
We therefore decompose \(R^{-}\) by textual role.
\(E^{-}\subset R^{-}\) contains task-relevant
observations, quantities, relations, and deductions supporting
\(y^{-}\), as highlighted in Fig.~\ref{fig:intro-recompute-reuse};
\(A^{-}\) is the explicit final-answer span, with
\(E^{-}\cap A^{-}=\varnothing\); the remainder is non-evidence context.
We compare the intact prior (N), evidence removal (E), the
length-matched non-evidence context control (C), and answer-span removal
(A), holding the model, image, question and prompt fixed. For conditions
\(X\) and \(Y\), we report
\(\Delta m_{\mathrm{X-Y}}=m(S_X)-m(S_Y)\), with analogous VUR and PAR
contrasts.

Tab.~\ref{tab:evidence-content} shows that removing \(E^{-}\) shifts
the answer-preference margin by \(2.069\) relative to N and by \(2.010\)
relative to C, with a positive E-minus-C effect in all \(16/16\)
models; C alone shifts the margin by only \(0.059\). Removing
\(A^{-}\) instead yields a smaller model-equal shift of \(0.074\), with
heterogeneous model-level effects. The dependence is also graded: as evidence removal
increases from \(0\%\) to \(100\%\), the proportion of samples with
\(m(S_{\alpha})>0\) increases monotonically at the model level in all
16 models (Fig.~\ref{fig:evidence-dose-response}).

Free generation exposes the same content specificity. Relative to C,
removing \(E^{-}\) raises VUR by \(43.76\) pp and lowers PAR by
\(62.58\) pp, with positive VUR effects in all \(16/16\) models;
relative to N, the changes are \(+45.22\) and \(-64.22\) pp. However,
the \emph{Other} rate also rises from \(8.35\%\) to \(26.50\%\)
(Tab.~\ref{tab:evidence-freegen-distribution}), so weakening stale
control does not by itself guarantee recovery. Across both metrics,
evidence-bearing content is the robust cross-model carrier of
textual-shortcut control, whereas answer-span effects remain
model-specific.

\noindent\textbf{Evidence Organization Modulates Shortcut Reuse.}
Content dependence alone does not distinguish isolated textual cues
from an organized reasoning path. In the coherent \(R^{-}\), visual
facts and deductions form an ordered route from premise to answer. We
reverse the evidence-bearing sentences and compare the resulting prior
(D) with the intact prior (N) and a matched non-evidence reorder (NE).

Disrupting evidence order weakens prior control. As shown in
Tab.~\ref{tab:evidence-organization}, it shifts the answer-preference
margin by \(+0.449\) relative to the intact prior, with positive effects
in \(14/16\) models. Relative to reordering only non-evidence
sentences, the disruption produces an additional \(+0.317\) margin
shift, again with positive effects in \(14/16\) models. This matched
comparison isolates sensitivity to the tested evidence ordering from
generic sentence reordering. The weakening also appears in free
generation: VUR increases by \(5.84\) pp and PAR decreases by
\(7.55\) pp.

\noindent\textbf{From Prior Influence to Textual Shortcut.}
Together, these interventions narrow broad prior-context influence to
an evidence-bearing path whose control scales with retained content and
is sensitive to the tested ordering. We call this organized path a
\emph{textual shortcut}: a reusable route to an answer without
re-deriving the relevant premises from the current image. We next ask
whether correcting the answer retires this shortcut.

\begin{table}[t]
\centering
\small
\setlength{\tabcolsep}{3pt}
\renewcommand{\arraystretch}{1.12}
\begin{tabular}{@{}lcc@{}}
\toprule
\multicolumn{1}{c}{}
& \multicolumn{2}{c}{\textbf{Free-generation prior-answer rate (\%)}} \\
\cmidrule(l){2-3}
\textbf{Evidence condition}
& \shortstack{\textbf{Direct}\\\textbf{inference }\(\mathbf{(D)}\)}
& \shortstack{\textbf{Prior-conditioned}\\\textbf{recovered }\(\mathbf{(H)}\)} \\
\midrule
\multicolumn{3}{@{}l}{\textit{Qwen3.5-27B}} \\
Full support \((F)\) & 1.0  & 10.6 \\
Withdrawn \((W)\)   & 4.0  & 84.8 \\
\addlinespace[2pt]
\multicolumn{3}{r@{}}{\textbf{Extra Old Lift \(=+71.2\) pp}} \\
\midrule
\multicolumn{3}{@{}l}{\textit{InternVL2-8B}} \\
Full support \((F)\) & 2.0  & 13.3 \\
Withdrawn \((W)\)   & 5.1  & 49.0 \\
\addlinespace[2pt]
\multicolumn{3}{r@{}}{\textbf{Extra Old Lift \(=+32.7\) pp}} \\
\midrule
\multicolumn{3}{@{}l}{\textit{Aggregate: all 16 VLMs}} \\
Full support \((F)\) & 1.34 & 9.81 \\
Withdrawn \((W)\)   & 7.17 & 69.01 \\
\addlinespace[2pt]
\multicolumn{3}{r@{}}{\textbf{Extra Old Lift \(=+53.38\) pp}} \\
\bottomrule
\end{tabular}
\caption{\textbf{Matched support withdrawal reveals prior-history
dependence after answer recovery.}
Combining rows \(F/W\) with columns \(D/H\) gives
the four states in Eq.~\ref{eq:extra-old-lift}.
}
\label{tab:shortcut-retirement}
\vspace{-1.2em}
\end{table}

\section{Correctness Is Not Shortcut Retirement}
\label{sec:shortcut-retirement}

A correct response does not distinguish shortcut retirement from
temporary override by current evidence. We separate these accounts by
stress-testing recovered responses under
matched support withdrawal, then restoring intact evidence to inspect
subsequent computation.

\noindent\textbf{Matched Support Withdrawal Reveals Residual Dependence.}
If recovery indicates shortcut retirement, weakening current support
should destabilize H no more than D. We test only \emph{recovered} pairs:
the prior-conditioned path (H) has produced \(y\) despite \(R^{-}\),
and its matched direct path (D) also produces \(y\) without \(R^{-}\).
For each path, we preserve the image and question, retain full current
support (F) or remove its evidence-bearing spans (W), then continue
generation and measure PAR. Because withdrawal may destabilize either
path, D controls for this generic effect. Let
\(S_H^F,S_H^W\) denote the full and withdrawn history contexts, with
\(S_D^F,S_D^W\) defined analogously. We measure the additional
resurgence associated with prior-conditioned history by
\begin{equation}
\begin{array}{rcl}
\mathrm{Extra\ Old\ Lift}
&=&
\left[\mathrm{PAR}(S_H^W)-\mathrm{PAR}(S_H^F)\right]
\\
&&-
\left[\mathrm{PAR}(S_D^W)-\mathrm{PAR}(S_D^F)\right]
\end{array}
\label{eq:extra-old-lift}
\end{equation}
A positive Extra Old Lift means that support withdrawal increases
prior-answer resurgence more on H than on the matched direct control.

\begin{figure}[t]
\centering
\includegraphics[width=\linewidth]{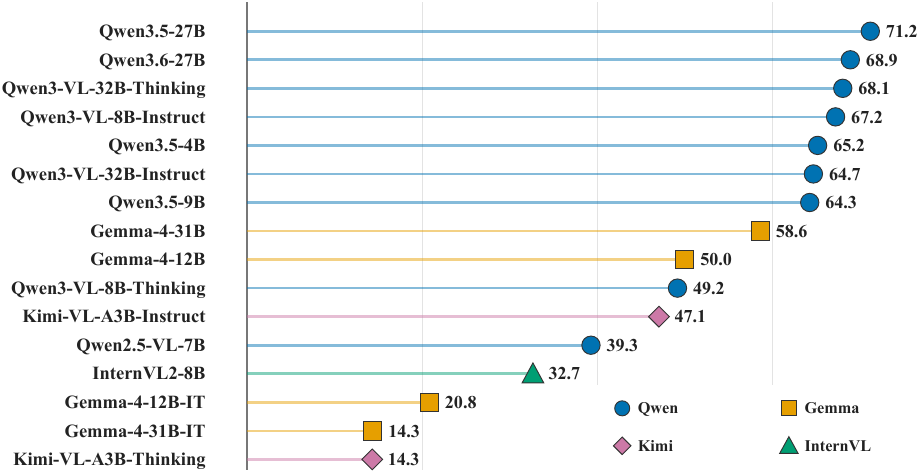}
\caption{\textbf{Extra Old Lift is positive across all 16
VLMs under matched support withdrawal.}
}
\label{fig:shortcut-retirement-models}
\vspace{-1.2em}
\end{figure}


Tab.~\ref{tab:shortcut-retirement} and
Fig.~\ref{fig:shortcut-retirement-models} show a consistent interaction.
In aggregate, PAR rises from \(9.81\%\) to \(69.01\%\) on H but only
from \(1.34\%\) to \(7.17\%\) on D, yielding an Extra Old Lift of
\(53.38\) pp. The model-equal effect is \(49.74\) pp, and all \(16/16\)
models are positive. Thus, a correct reflected answer can coexist with
detectable prior-history dependence, consistent with current support
overriding rather than eliminating the shortcut. This stress test does
not imply spontaneous relapse under intact support, which we test next.

\noindent\textbf{Residual Prior Influence Under Intact Support.}
Support withdrawal reveals susceptibility, but not its expression
under intact evidence. We therefore retain \(I\) and the complete
trajectory, append either an equivalent re-query or a derived
computation to D and H, and stratify H by its initial outcome:
\emph{recovered} or \emph{active-stale}. This design distinguishes
answer recurrence from premise transfer across observed states.
Tab.~\ref{tab:future-state-continuation} reports H-minus-D effects.
Negative \(\Delta m\) favors the prior; negative
\(\Delta\mathrm{VUR}\) denotes fewer current-target generations, and
positive \(\Delta\mathrm{PAR}\) more prior-target generations. For
derived computation, VUR and PAR refer to the current- and prior-derived
targets.

\noindent\emph{Equivalent re-query.}
We ask whether the prior answer recurs when \(Q\) is posed again under
the same image; prompt variants appear in
App.~\ref{app:equivalent-requery}. For recovered outcomes, history
shifts the candidate margin priorward by \(-0.547\), but free-generation
changes are modest: VUR decreases by \(2.14\) pp and PAR increases by
\(0.98\) pp. The score shift therefore rarely crosses the generation
boundary. For active-stale outcomes, the shift is much larger
(\(\Delta m=-3.896\)): VUR decreases by \(39.88\) pp and PAR increases
by \(35.05\) pp. Equivalent re-query thus reveals weak answer recurrence
after recovery but strong recurrence while the shortcut remains active.

\noindent\emph{Derived computation.}
We ask whether a stale premise transfers when a new operation is
applied to the task-relevant quantity; construction details appear in
App.~\ref{app:derived-computation}. For recovered outcomes,
\(\Delta m=-0.391\) and current-derived generation falls by \(18.99\)
pp, but prior-derived generation changes by only \(+0.07\) pp. History
therefore disrupts the new computation without reliably substituting
the prior premise. For active-stale outcomes, \(\Delta m=-3.756\),
current-derived generation falls by \(65.19\) pp, and prior-derived
generation rises by \(45.28\) pp, showing that an active shortcut can
transfer a stale premise into a new operation. The larger recovered VUR
loss is consistent with derived computation requiring premise selection
plus a new operation, whereas re-query can reuse the current-consistent
answer.

\noindent\textbf{Residual Influence Depends on State and Continuation.}
Together, the results show that answer correction does not certify
shortcut retirement. Withdrawal exposes prior-history dependence after
recovery, whereas intact-support effects are state- and
continuation-dependent: recovered outcomes show weak recurrence or
computational disruption without reliable stale-premise substitution,
while active-stale outcomes show strong recurrence and premise
transfer.

\begin{table}[t]
\centering
\small
\setlength{\tabcolsep}{2.9pt}
\renewcommand{\arraystretch}{1.10}
\begin{tabular}{@{}llrc@{}}
\toprule
\textbf{Continuation}
& \shortstack{\textbf{Initial outcome}}
& \(\Delta m_{\mathrm{H-D}}\)
& \shortstack{
\(\Delta\mathrm{VUR}\) / \(\Delta\mathrm{PAR}\)
} \\
\midrule
\multicolumn{4}{@{}l}{\textit{Qwen3-VL-8B (Instruct)}} \\
\multirow{2}{*}{\shortstack{Equivalent \\ Re-query}}
& \shortstack[l]{Recovered\\[-1pt]{\scriptsize (70.05\%)}}
& -0.707
& -1.72 / +0.34 \\
& \shortstack[l]{Active stale\\[-1pt]{\scriptsize (25.36\%)}}
& -7.759
& -47.14 / +37.62 \\
\midrule
\multirow{2}{*}{\shortstack{Derived \\ Computation}}
& \shortstack[l]{Recovered\\[-1pt]{\scriptsize (71.85\%)}}
& -0.562
& -11.11 / +0.19 \\
& \shortstack[l]{Active stale\\[-1pt]{\scriptsize (23.95\%)}}
& -7.279
& -82.46 / +66.67 \\
\midrule
\midrule
\multicolumn{4}{@{}l}{\textit{Kimi-VL-A3B (Instruct)}} \\
\multirow{2}{*}{\shortstack{Equivalent \\ Re-query}}
& \shortstack[l]{Recovered\\[-1pt]{\scriptsize (28.18\%)}}
& -0.904
& -8.82 / +2.94 \\
& \shortstack[l]{Active stale\\[-1pt]{\scriptsize (66.30\%)}}
& -1.854
& -57.08 / +47.92 \\
\midrule
\multirow{2}{*}{\shortstack{Derived \\ Computation}}
& \shortstack[l]{Recovered\\[-1pt]{\scriptsize (30.48\%)}}
& -0.320
& -10.42 / +1.04 \\
& \shortstack[l]{Active stale\\[-1pt]{\scriptsize (63.81\%)}}
& -2.192
& -56.72 / +49.75 \\
\midrule
\midrule
\multicolumn{4}{@{}l}{\textit{Aggregate: all 16 VLMs}} \\
\multirow{2}{*}{\shortstack{Equivalent \\ Re-query}}
& \shortstack[l]{Recovered\\[-1pt]{\scriptsize (59.05\%)}}
& \textbf{-0.547}
& -2.14 / +0.98 \\
& \shortstack[l]{Active stale\\[-1pt]{\scriptsize (34.09\%)}}
& -3.896
& -39.88 / \textbf{+35.05} \\
\midrule
\multirow{2}{*}{\shortstack{Derived \\ Computation}}
& \shortstack[l]{Recovered\\[-1pt]{\scriptsize (60.82\%)}}
& -0.391
& -18.99 / +0.07 \\
& \shortstack[l]{Active stale\\[-1pt]{\scriptsize (32.30\%)}}
& -3.756
& -65.19 / \textbf{+45.28} \\
\bottomrule
\end{tabular}
\caption{\textbf{Prior influence depends on initial outcome and
continuation.}
H--D denotes prior-conditioned minus direct. Negative \(\Delta m\)
favors the prior; \(\Delta\mathrm{VUR}/\Delta\mathrm{PAR}\) (pp) refer
to current/prior targets for re-query and current-/prior-derived targets
for derived computation. Parentheses show outcome shares; bold marks
aggregate effects.}
\label{tab:future-state-continuation}
\vspace{-1.1em}
\end{table}

\begin{table}[t]
\centering
\small
\setlength{\tabcolsep}{1.5pt}
\renewcommand{\arraystretch}{1.08}
\begin{tabular*}{\columnwidth}{@{\extracolsep{\fill}}lcccc@{}}
\toprule
\multirow{2}{*}{\textbf{Metric}}
& \multicolumn{2}{c}{\textbf{Qwen3-VL-8B}}
& \multicolumn{2}{c}{\textbf{Qwen3-VL-32B}} \\
\cmidrule(lr){2-3}\cmidrule(lr){4-5}
& \textbf{Instruct} & \textbf{Thinking}
& \textbf{Instruct} & \textbf{Thinking} \\
\midrule
\multicolumn{5}{@{}l}{\textit{Prompt paraphrases (mean \(\pm\) std.)}} \\
VUR (\%) \(\uparrow\) & 49.1 \(\pm\) 2.8 & 16.3 \(\pm\) 2.1 & 58.7 \(\pm\) 2.6 & 32.2 \(\pm\) 3.1 \\
PAR (\%) \(\downarrow\) & 28.0 \(\pm\) 2.4 & 55.8 \(\pm\) 3.0 & 20.1 \(\pm\) 2.2 & 46.8 \(\pm\) 2.7 \\
\midrule
\multicolumn{5}{@{}l}{\textit{Prefix control (reflection / neutral)}} \\
VUR (\%) \(\uparrow\) & 49.5 / 48.8 & 15.8 / 16.4 & 59.0 / 58.3 & 31.8 / 32.5 \\
PAR (\%) \(\downarrow\) & 27.6 / 28.2 & 56.2 / 55.4 & 19.6 / 20.1 & 47.2 / 46.6 \\
\midrule
\multicolumn{5}{@{}l}{\textit{Naturally occurring errors (reflection / prompt-only)}} \\
\shortstack[l]{Correction \\ Rate (\%)} \(\uparrow\) & 34.7 / 52.9 & 20.8 / 46.1 & 42.3 / 61.4 & 25.6 / 52.0 \\
\bottomrule
\end{tabular*}
\caption{\textbf{Robustness to prompt wording, prefix semantics, and
prior construction.}}
\label{tab:prompt-prefix-robustness}
\vspace{-0.8em}
\end{table}

\noindent\textbf{Robustness Beyond the Controlled Construction.}
Tab.~\ref{tab:prompt-prefix-robustness} tests whether the effect depends on prompt wording, reflection-prefix semantics, or counterfactual construction.

\noindent\textit{Prompt wording.}
To test whether prior-CoT influence is an artifact of fixed wording,
we evaluate ten semantically matched reflection-prompt paraphrases.
The first block of Tab.~\ref{tab:prompt-prefix-robustness} reports the
mean and standard deviation across the ten variants listed in
App.~\ref{app:prompt-robustness},
Tab.~\ref{tab:reflection-prompt-variants}. Across the four
Qwen3-VL variants, the standard deviations are at most \(3.1\) pp for
VUR and \(3.0\) pp for PAR, showing that the observed behavior is
consistent across prompt phrasings.

\noindent\textit{Reflection prefix.}
To isolate reflection-specific prefix semantics, we replace the
reflection prefix with a neutral continuation while holding all other
inputs and settings fixed; entries are ordered as reflection / neutral.
VUR and PAR change by at most \(0.8\) pp, indicating that the effect is
not induced by the reflective wording itself.

\noindent\textit{Naturally occurring errors.}
To test dependence on counterfactual prior
construction, we collect initial errors produced during standard
inference on the original benchmarks and report reflection /
prompt-only correction rates. Retaining the erroneous CoT reduces
correction rates by \(18.2\)--\(26.4\) pp relative to the matched
prompt-only condition, showing that prior-CoT influence also appears
without an image swap.

\section{Fresh-State Attention Firewall}
\label{sec:fsaf}

\noindent\textbf{Motivation.}
Previous results identify \(R^{-}\) as a reusable
evidence-bearing shortcut and show that a corrected answer does not
certify that its influence has been removed. We therefore restrict the
accessibility of \(R^{-}\), rather than merely prompting the model to
inspect \(I\) again. \emph{Fresh-State Attention Firewall} (FSAF) is a
training-free, single-pass intervention with two coupled components: a
new user--assistant boundary that starts a fresh re-examination, and an
attention mask that isolates this fresh suffix from \(R^{-}\) during
both prefill and generation. It retains the original conversation and
requires neither target-answer access nor an evidence parser.

\noindent\textbf{Fresh re-query boundary.}
Given the prior-conditioned context containing \(R^{-}\), FSAF appends
a fixed user re-query \(U_f\) that asks the model to answer \(Q\) again
using only the current image \(I\). Let \(R^{F}\) denote the following
assistant response and \(\mathcal{F}=U_f\cup R^{F}\) the fresh suffix.
The new boundary separates this re-examination from the previous
assistant trace, but does not by itself isolate their computations:
during prefill, tokens in \(U_f\) can still attend to \(R^{-}\) and
carry its content into generation through their cached states.

\noindent\textbf{Suffix attention firewall.}
The firewall closes both this indirect route and direct access from the
new response. At every language layer, FSAF augments the standard
Transformer attention logits \citep{vaswani2017attention} with the
following mask:
\begin{equation}
M^{\mathrm{FSAF}}_{qk}
=
\left\{
\begin{array}{ll}
-\infty, & q\in\mathcal{F}\ \mathrm{and}\ k\in R^{-}\\
0, & \mathrm{otherwise}
\end{array}
\right.
\label{eq:fsaf-mask}
\end{equation}
\begin{equation}
A^{(\ell)}_{q,:}
=
\mathrm{softmax}\!\left(
\mathbf{L}^{(\ell)}_{q,:}
+M^{\mathrm{causal}}_{q,:}
+M^{\mathrm{FSAF}}_{q,:}
\right)
\label{eq:fsaf-attention}
\end{equation}
where \(\mathbf{L}^{(\ell)}\) denotes the matrix of unmasked attention
logits at layer \(\ell\). Applying the mask while prefilling \(U_f\)
prevents its cached states from absorbing \(R^{-}\); keeping it active
during autoregressive generation prevents tokens in \(R^{F}\) from
reading \(R^{-}\) directly. All other causal edges remain unchanged, so
the fresh suffix can still access \(I\), \(Q\), system tokens, \(U_f\),
and earlier tokens in \(R^{F}\). The softmax renormalizes over these
remaining keys: FSAF removes the stale textual path without deleting
the transcript or prescribing a fixed increase in visual attention.
The exact re-query, execution procedure, implementation, and
design ablations are provided in App.~\ref{app:fsaf-details}.

\begin{table}[t]
\centering
\small
\setlength{\tabcolsep}{1.0pt}
\renewcommand{\arraystretch}{1.08}
\begin{tabular*}{\columnwidth}{@{\extracolsep{\fill}}lccc@{}}
\toprule
\textbf{Model}
& \textbf{VUR (\%) \(\uparrow\)}
& \textbf{PAR (\%) \(\downarrow\)}
& \textbf{\(m\) \(\uparrow\)}\\
\midrule
\shortstack[l]{Qwen2.5-\\VL-7B (I)}
& \shortstack{23.60 \(\rightarrow\) 42.32\\(+18.73)}
& \shortstack{41.20 \(\rightarrow\) 9.74\\(-31.46)}
& \shortstack{-1.75 \(\rightarrow\) +1.52\\(+3.26)}\\
\midrule
\shortstack[l]{Qwen3-\\VL-8B (I)}
& \shortstack{49.52 \(\rightarrow\) 65.48\\(+15.95)}
& \shortstack{27.62 \(\rightarrow\) 5.95\\(-21.67)}
& \shortstack{-5.64 \(\rightarrow\) +3.59\\(+9.23)}\\
\midrule
\shortstack[l]{Qwen3-\\VL-8B (T)}
& \shortstack{15.84 \(\rightarrow\) 35.80\\(+19.96)}
& \shortstack{56.17 \(\rightarrow\) 1.03\\(-55.14)}
& \shortstack{-6.01 \(\rightarrow\) +1.37\\(+7.38)}\\
\midrule
\shortstack[l]{Qwen3-\\VL-32B (I)}
& \shortstack{59.01 \(\rightarrow\) 74.22\\(+15.22)}
& \shortstack{19.57 \(\rightarrow\) 1.55\\(-18.01)}
& \shortstack{-7.52 \(\rightarrow\) +4.76\\(+12.27)}\\
\midrule
\shortstack[l]{Qwen3-\\VL-32B (T)}
& \shortstack{31.80 \(\rightarrow\) 53.77\\(+21.97)}
& \shortstack{47.21 \(\rightarrow\) 1.64\\(-45.57)}
& \shortstack{-5.62 \(\rightarrow\) +4.03\\(+9.65)}\\
\midrule
\shortstack[l]{\textbf{Average}\\\strut}
& \shortstack{\textbf{35.28} \(\rightarrow\) \textbf{53.61}\\\textbf{(+18.33)}}
& \shortstack{\textbf{39.22} \(\rightarrow\) \textbf{3.67}\\\textbf{(-35.56)}}
& \shortstack{\textbf{-5.49} \(\rightarrow\) \textbf{+2.97}\\\textbf{(+8.46)}}\\
\bottomrule
\end{tabular*}
\caption{\textbf{FSAF improves visual updating while suppressing prior-answer retention.}
Each cell reports Baseline \(\rightarrow\) FSAF, with the paired change
in parentheses.
}
\label{tab:fsaf-main}
\vspace{-0.6em}
\end{table}


\noindent\textbf{Evaluation.}
To exclude failures caused by task-solving difficulty, we evaluate FSAF
on  VS-Bench model--sample pairs from five Qwen models
for which each model correctly answers both the current and
counterfactual images under direct inference. For each pair, we compare
ordinary prior-conditioned reflection with FSAF under identical settings.

\noindent\textbf{Results.}
In Tab.~\ref{tab:fsaf-main}, FSAF improves all three readouts for every evaluated model. Pooled
across model--sample pairs, it raises VUR from \(35.28\%\) to
\(53.61\%\), reduces PAR from \(39.22\%\) to \(3.67\%\), and shifts the
answer-preference margin from \(-5.49\) to \(+2.97\).
Together, these results show that FSAF consistently reduces stale
prior-CoT influence across all evaluated Qwen models.
FSAF suppresses stale-answer
control without guaranteeing current-answer recovery on every pair.

\section{Conclusion}


This work asks what VLMs reuse when self-reflection fails to recompute
after visual evidence changes. Across 16 VLMs, controlled removal and
reordering identify evidence-bearing prior reasoning as the most robust
carrier of textual-shortcut control. A correct answer, however, does
not certify shortcut retirement.
Residual dependence can remain after recovery, whereas strong answer
recurrence and stale-premise transfer are concentrated when the
shortcut remains active. Guided by this diagnosis, FSAF blocks fresh
computation from attending to the prior CoT and improves visual updating
across five Qwen VLMs. Reliable VLM self-reflection therefore requires
more than looking again: fresh visual recomputation must be protected
from stale textual reuse.

\clearpage
\bibliography{arxiv,related_work}

@inproceedings{madaan2023selfrefine,
  author    = {Aman Madaan and Niket Tandon and Prakhar Gupta and Skyler Hallinan and Luyu Gao and Sarah Wiegreffe and Uri Alon and Nouha Dziri and Shrimai Prabhumoye and Yiming Yang and Shashank Gupta and Bodhisattwa Prasad Majumder and Katherine Hermann and Sean Welleck and Amir Yazdanbakhsh and Peter Clark},
  title     = {Self-Refine: Iterative Refinement with Self-Feedback},
  booktitle = {Advances in Neural Information Processing Systems},
  volume    = {36},
  year      = {2023},
  url       = {https://papers.nips.cc/paper_files/paper/2023/hash/91edff07232fb1b55a505a9e9f6c0ff3-Abstract-Conference.html}
}

@article{kamoi2024when,
  author    = {Ryo Kamoi and Yusen Zhang and Nan Zhang and Jiawei Han and Rui Zhang},
  title     = {When Can {LLM}s Actually Correct Their Own Mistakes? A Critical Survey of Self-Correction of {LLM}s},
  journal   = {Transactions of the Association for Computational Linguistics},
  volume    = {12},
  pages     = {1417--1440},
  year      = {2024},
  publisher = {MIT Press},
  doi       = {10.1162/tacl_a_00713},
  url       = {https://aclanthology.org/2024.tacl-1.78/}
}

@misc{shi2026visualswap,
  author        = {Chufan Shi and Cheng Yang and Yaokang Wu and Linghao Jin and Bo Shui and Taylor Berg-Kirkpatrick and Xuezhe Ma},
  title         = {Are {VLM}s Seeing or Just Saying? Uncovering the Illusion of Visual Re-examination},
  year          = {2026},
  eprint        = {2605.15864},
  archivePrefix = {arXiv},
  primaryClass  = {cs.CV},
  doi           = {10.48550/arXiv.2605.15864},
  url           = {https://arxiv.org/abs/2605.15864},
  note          = {Accepted as an oral presentation at ICML 2026}
}

@inproceedings{xu2025morethinking,
  author    = {Zhongxing Xu and Chengzhi Liu and Qingyue Wei and Juncheng Wu and James Y. Zou and Xin Wang and Yuyin Zhou and Sheng Liu},
  title     = {More Thinking, Less Seeing? Assessing Amplified Hallucination in Multimodal Reasoning Models},
  booktitle = {Advances in Neural Information Processing Systems},
  volume    = {38},
  year      = {2025},
  url       = {https://papers.nips.cc/paper_files/paper/2025/hash/777db387a5ccb131ba8c7cd155166b85-Abstract-Conference.html}
}

@inproceedings{jian2025lookagain,
  author    = {Pu Jian and Junhong Wu and Wei Sun and Chen Wang and Shuo Ren and Jiajun Zhang},
  title     = {Look Again, Think Slowly: Enhancing Visual Reflection in Vision-Language Models},
  booktitle = {Proceedings of the 2025 Conference on Empirical Methods in Natural Language Processing},
  pages     = {9251--9270},
  year      = {2025},
  address   = {Suzhou, China},
  publisher = {Association for Computational Linguistics},
  doi       = {10.18653/v1/2025.emnlp-main.470},
  url       = {https://aclanthology.org/2025.emnlp-main.470/}
}

@article{yang2026lookback,
  author    = {Shuo Yang and Yuwei Niu and Yuyang Liu and Yang Ye and Bin Lin and Li Yuan},
  title     = {Look-Back: Implicit Visual Re-focusing in {MLLM} Reasoning},
  journal   = {Proceedings of the AAAI Conference on Artificial Intelligence},
  volume    = {40},
  number    = {14},
  pages     = {11694--11702},
  year      = {2026},
  doi       = {10.1609/aaai.v40i14.38154},
  url       = {https://ojs.aaai.org/index.php/AAAI/article/view/38154}
}

@inproceedings{huang2024cannot,
  author    = {Jie Huang and Xinyun Chen and Swaroop Mishra and Huaixiu Steven Zheng and Adams Wei Yu and Xinying Song and Denny Zhou},
  title     = {Large Language Models Cannot Self-Correct Reasoning Yet},
  booktitle = {The Twelfth International Conference on Learning Representations},
  year      = {2024},
  publisher = {OpenReview.net},
  url       = {https://openreview.net/forum?id=IkmD3fKBPQ}
}

@inproceedings{wilie2024beliefrevision,
  author    = {Bryan Wilie and Samuel Cahyawijaya and Etsuko Ishii and Junxian He and Pascale Fung},
  title     = {Belief Revision: The Adaptability of Large Language Models Reasoning},
  booktitle = {Proceedings of the 2024 Conference on Empirical Methods in Natural Language Processing},
  pages     = {10480--10496},
  year      = {2024},
  address   = {Miami, Florida, USA},
  publisher = {Association for Computational Linguistics},
  doi       = {10.18653/v1/2024.emnlp-main.586},
  url       = {https://aclanthology.org/2024.emnlp-main.586/}
}

@inproceedings{malladi2023finetuning,
  title={Fine-Tuning Language Models with Just Forward Passes},
  author={Malladi, Sadhika and Gao, Tianyu and Nichani, Eshaan
          and Damian, Alex and Lee, Jason D. and Chen, Danqi
          and Arora, Sanjeev},
  booktitle={Advances in Neural Information Processing Systems},
  volume={36},
  year={2023}
}

@inproceedings{ren2023selfevaluation,
  author       = {Jie Ren and
                  Yao Zhao and
                  Tu Vu and
                  Peter J. Liu and
                  Balaji Lakshminarayanan},
  editor       = {Javier Antor{\'{a}}n and
                  Arno Blaas and
                  Kelly Buchanan and
                  Fan Feng and
                  Vincent Fortuin and
                  Sahra Ghalebikesabi and
                  Andreas Kriegler and
                  Ian Mason and
                  David Rohde and
                  Francisco J. R. Ruiz and
                  Tobias Uelwer and
                  Yubin Xie and
                  Rui Yang},
  title        = {Self-Evaluation Improves Selective Generation in Large Language Models},
  booktitle    = {Proceedings on "I Can't Believe It's Not Better: Failure Modes in
                  the Age of Foundation Models" at NeurIPS 2023 Workshops, 16 December
                  2023, New Orleans, Louisiana, {USA}},
  series       = {Proceedings of Machine Learning Research},
  volume       = {239},
  pages        = {49--64},
  publisher    = {{PMLR}},
  year         = {2023},
  url          = {https://proceedings.mlr.press/v239/ren23a.html},
  bibsource    = {dblp computer science bibliography, https://dblp.org}
}

@inproceedings{li2024multiplechoice,
  title={Can Multiple-choice Questions Really Be Useful
         in Detecting the Abilities of {LLM}s?},
  author={Li, Wangyue and Li, Liangzhi and Xiang, Tong and Liu, Xiao
          and Deng, Wei and Garcia, Noa},
  booktitle={Proceedings of the 2024 Joint International Conference
             on Computational Linguistics, Language Resources and Evaluation},
  pages={2819--2834},
  year={2024}
}

@misc{chu2025qwenlookagain,
  author        = {Xu Chu and Xinrong Chen and Guanyu Wang and Zhijie Tan and Kui Huang and Wenyu Lv and Tong Mo and Weiping Li},
  title         = {Qwen Look Again: Guiding Vision-Language Reasoning Models to Re-attention Visual Information},
  year          = {2025},
  eprint        = {2505.23558},
  archivePrefix = {arXiv},
  primaryClass  = {cs.CV},
  doi           = {10.48550/arXiv.2505.23558},
  url           = {https://arxiv.org/abs/2505.23558}
}

@inproceedings{cheng2025r3v,
  author    = {Kanzhi Cheng and YanTao, Li and Fangzhi Xu and Jianbing Zhang and Hao Zhou and Yang Liu},
  title     = {Vision-Language Models Can Self-Improve Reasoning via Reflection},
  booktitle = {Proceedings of the 2025 Conference of the Nations of the Americas Chapter of the Association for Computational Linguistics: Human Language Technologies (Volume 1: Long Papers)},
  pages     = {8876--8892},
  year      = {2025},
  publisher = {Association for Computational Linguistics},
  doi       = {10.18653/v1/2025.naacl-long.447},
  url       = {https://aclanthology.org/2025.naacl-long.447/}
}

@inproceedings{ji2025reflair,
  author    = {Jiazhou Ji and Xinru Lu},
  title     = {{ReFLAIR}: Enhancing Multimodal Reasoning via Structured Reflection and Reward-Guided Learning},
  booktitle = {Findings of the Association for Computational Linguistics: EMNLP 2025},
  pages     = {25401--25413},
  year      = {2025},
  publisher = {Association for Computational Linguistics},
  doi       = {10.18653/v1/2025.findings-emnlp.1384},
  url       = {https://aclanthology.org/2025.findings-emnlp.1384/}
}

@inproceedings{uppaal2026journey,
  author    = {Rheeya Uppaal and Phu Mon Htut and Min Bai and Nikolaos Pappas and Zheng Qi and Sandesh Swamy},
  title     = {Journey Before Destination: On the importance of Visual Faithfulness in Slow Thinking},
  booktitle = {Proceedings of the 19th Conference of the European Chapter of the Association for Computational Linguistics (Volume 1: Long Papers)},
  pages     = {4147--4168},
  year      = {2026},
  publisher = {Association for Computational Linguistics},
  doi       = {10.18653/v1/2026.eacl-long.194},
  url       = {https://aclanthology.org/2026.eacl-long.194/}
}

@misc{tang2026groundedreflection,
  author        = {Liyan Tang and Fangcong Yin and Greg Durrett},
  title         = {Visually Grounded Self-Reflection for Vision-Language Models via Reinforcement Learning},
  year          = {2026},
  eprint        = {2607.02490},
  archivePrefix = {arXiv},
  primaryClass  = {cs.CL},
  doi           = {10.48550/arXiv.2607.02490},
  url           = {https://arxiv.org/abs/2607.02490}
}

@inproceedings{sun2026perceive,
  author    = {Xiaoxiao Sun and Mingyang Li and Kun Yuan and Min Woo Sun and Mark Endo and Shengguang Wu and Changlin Li and Yuhui Zhang and Zeyu Wang and Serena Yeung-Levy},
  title     = {Do {VLM}s Perceive or Recall? Probing Visual Perception vs. Memory with Classic Visual Illusions},
  booktitle = {Proceedings of the IEEE/CVF Conference on Computer Vision and Pattern Recognition (CVPR)},
  pages     = {25861--25870},
  year      = {2026},
  url       = {https://openaccess.thecvf.com/content/CVPR2026/html/Sun_Do_VLMs_Perceive_or_Recall_Probing_Visual_Perception_vs._Memory_CVPR_2026_paper.html}
}

@inproceedings{shinn2023reflexion,
  author    = {Noah Shinn and Federico Cassano and Ashwin Gopinath and Karthik Narasimhan and Shunyu Yao},
  title     = {Reflexion: Language Agents with Verbal Reinforcement Learning},
  booktitle = {Advances in Neural Information Processing Systems},
  volume    = {36},
  year      = {2023},
  url       = {https://papers.neurips.cc/paper_files/paper/2023/hash/1b44b878bb782e6954cd888628510e90-Abstract-Conference.html}
}

@inproceedings{gou2024critic,
  author    = {Zhibin Gou and Zhihong Shao and Yeyun Gong and Yelong Shen and Yujiu Yang and Nan Duan and Weizhu Chen},
  title     = {{CRITIC}: Large Language Models Can Self-Correct with Tool-Interactive Critiquing},
  booktitle = {The Twelfth International Conference on Learning Representations},
  year      = {2024},
  url       = {https://openreview.net/forum?id=WSrRF5Wy6v}
}

@article{pan2024automatic,
  author    = {Liangming Pan and Michael Saxon and Wenda Xu and Deepak Nathani and Xinyi Wang and William Yang Wang},
  title     = {Automatically Correcting Large Language Models: Surveying the Landscape of Diverse Automated Correction Strategies},
  journal   = {Transactions of the Association for Computational Linguistics},
  volume    = {12},
  pages     = {484--506},
  year      = {2024},
  doi       = {10.1162/tacl_a_00660},
  url       = {https://aclanthology.org/2024.tacl-1.27/}
}

@inproceedings{zhang2024selfcontrast,
  author    = {Wenqi Zhang and Yongliang Shen and Linjuan Wu and Qiuying Peng and Jun Wang and Yueting Zhuang and Weiming Lu},
  title     = {Self-Contrast: Better Reflection Through Inconsistent Solving Perspectives},
  booktitle = {Proceedings of the 62nd Annual Meeting of the Association for Computational Linguistics (Volume 1: Long Papers)},
  pages     = {3602--3622},
  year      = {2024},
  publisher = {Association for Computational Linguistics},
  doi       = {10.18653/v1/2024.acl-long.197},
  url       = {https://aclanthology.org/2024.acl-long.197/}
}

@inproceedings{yang2025confidence,
  author    = {Zhe Yang and Yichang Zhang and Yudong Wang and Ziyao Xu and Junyang Lin and Zhifang Sui},
  title     = {Confidence v.s. Critique: A Decomposition of Self-Correction Capability for {LLM}s},
  booktitle = {Proceedings of the 63rd Annual Meeting of the Association for Computational Linguistics (Volume 1: Long Papers)},
  pages     = {3998--4014},
  year      = {2025},
  publisher = {Association for Computational Linguistics},
  doi       = {10.18653/v1/2025.acl-long.203},
  url       = {https://aclanthology.org/2025.acl-long.203/}
}

@inproceedings{zhang2024darkside,
    title = "Understanding the Dark Side of {LLM}s' Intrinsic Self-Correction",
    author = "Zhang, Qingjie  and
      Wang, Di  and
      Qian, Haoting  and
      Li, Yiming  and
      Zhang, Tianwei  and
      Huang, Minlie  and
      Xu, Ke  and
      Li, Hewu  and
      Yan, Liu  and
      Qiu, Han",
    editor = "Che, Wanxiang  and
      Nabende, Joyce  and
      Shutova, Ekaterina  and
      Pilehvar, Mohammad Taher",
    booktitle = "Proceedings of the 63rd Annual Meeting of the Association for Computational Linguistics (Volume 1: Long Papers)",
    month = jul,
    year = "2025",
    address = "Vienna, Austria",
    publisher = "Association for Computational Linguistics",
    url = "https://aclanthology.org/2025.acl-long.1314/",
    doi = "10.18653/v1/2025.acl-long.1314",
    pages = "27066--27101",
    ISBN = "979-8-89176-251-0"
}

@misc{simhi2026oldhabits,
  author        = {Adi Simhi and Fazl Barez and Martin Tutek and Yonatan Belinkov and Shay B. Cohen},
  title         = {Old Habits Die Hard: How Conversational History Geometrically Traps {LLM}s},
  year          = {2026},
  eprint        = {2603.03308},
  archivePrefix = {arXiv},
  primaryClass  = {cs.CL},
  doi           = {10.48550/arXiv.2603.03308},
  url           = {https://arxiv.org/abs/2603.03308}
}

@misc{chen2026selfcorrection,
  author        = {Kuan-Yen Chen and Fang-Yi Su and Jung-Hsien Chiang},
  title         = {The Self-Correction Illusion: {LLM}s Correct Others but Not Themselves},
  year          = {2026},
  eprint        = {2606.05976},
  archivePrefix = {arXiv},
  primaryClass  = {cs.CL},
  doi           = {10.48550/arXiv.2606.05976},
  url           = {https://arxiv.org/abs/2606.05976}
}

@inproceedings{pan2026userassistant,
  author    = {Xu Pan and Jingxuan Fan and Zidi Xiong and Ely Hahami and Jorin Overwiening and Ziqian Xie},
  title     = {User-Assistant Bias in {LLM}s},
  booktitle = {Findings of the Association for Computational Linguistics: ACL 2026},
  pages     = {9218--9241},
  year      = {2026},
  publisher = {Association for Computational Linguistics},
  doi       = {10.18653/v1/2026.findings-acl.449},
  url       = {https://aclanthology.org/2026.findings-acl.449/}
}

@inproceedings{turpin2023unfaithful,
  author    = {Miles Turpin and Julian Michael and Ethan Perez and Samuel R. Bowman},
  title     = {Language Models Don't Always Say What They Think: Unfaithful Explanations in Chain-of-Thought Prompting},
  booktitle = {Advances in Neural Information Processing Systems},
  volume    = {36},
  year      = {2023},
  url       = {https://proceedings.neurips.cc/paper_files/paper/2023/hash/ed3fea9033a80fea1376299fa7863f4a-Abstract-Conference.html}
}

@article{lanham2023faithfulness,
  author       = {Tamera Lanham and
                  Anna Chen and
                  Ansh Radhakrishnan and
                  Benoit Steiner and
                  Carson Denison and
                  Danny Hernandez and
                  Dustin Li and
                  Esin Durmus and
                  Evan Hubinger and
                  Jackson Kernion and
                  Kamile Lukosiute and
                  Karina Nguyen and
                  Newton Cheng and
                  Nicholas Joseph and
                  Nicholas Schiefer and
                  Oliver Rausch and
                  Robin Larson and
                  Sam McCandlish and
                  Sandipan Kundu and
                  Saurav Kadavath and
                  Shannon Yang and
                  Thomas Henighan and
                  Timothy Maxwell and
                  Timothy Telleen{-}Lawton and
                  Tristan Hume and
                  Zac Hatfield{-}Dodds and
                  Jared Kaplan and
                  Jan Brauner and
                  Samuel R. Bowman and
                  Ethan Perez},
  title        = {Measuring Faithfulness in Chain-of-Thought Reasoning},
  journal      = {CoRR},
  volume       = {abs/2307.13702},
  year         = {2023},
  url          = {https://doi.org/10.48550/arXiv.2307.13702},
  doi          = {10.48550/ARXIV.2307.13702},
  eprinttype   = {arXiv},
  eprint       = {2307.13702},
  bibsource    = {dblp computer science bibliography, https://dblp.org}
}

@inproceedings{lu2024mathvista,
  author    = {Pan Lu and Hritik Bansal and Tony Xia and Jiacheng Liu and Chunyuan Li and Hannaneh Hajishirzi and Hao Cheng and Kai-Wei Chang and Michel Galley and Jianfeng Gao},
  title     = {{MathVista}: Evaluating Mathematical Reasoning of Foundation Models in Visual Contexts},
  booktitle = {The Twelfth International Conference on Learning Representations},
  year      = {2024},
  url       = {https://proceedings.iclr.cc/paper_files/paper/2024/hash/663bce02a0050c4a11f1eb8a7f1429d3-Abstract-Conference.html}
}

@inproceedings{zhang2024mathverse,
  author       = {Renrui Zhang and
                  Dongzhi Jiang and
                  Yichi Zhang and
                  Haokun Lin and
                  Ziyu Guo and
                  Pengshuo Qiu and
                  Aojun Zhou and
                  Pan Lu and
                  Kai{-}Wei Chang and
                  Yu Qiao and
                  Peng Gao and
                  Hongsheng Li},
  editor       = {Ales Leonardis and
                  Elisa Ricci and
                  Stefan Roth and
                  Olga Russakovsky and
                  Torsten Sattler and
                  G{\"{u}}l Varol},
  title        = {{MATHVERSE:} Does Your Multi-modal {LLM} Truly See the Diagrams in
                  Visual Math Problems?},
  booktitle    = {Computer Vision - {ECCV} 2024 - 18th European Conference, Milan, Italy,
                  September 29-October 4, 2024, Proceedings, Part {VIII}},
  series       = {Lecture Notes in Computer Science},
  volume       = {15066},
  pages        = {169--186},
  publisher    = {Springer},
  year         = {2024},
  url          = {https://doi.org/10.1007/978-3-031-73242-3\_10},
  doi          = {10.1007/978-3-031-73242-3\_10},
  bibsource    = {dblp computer science bibliography, https://dblp.org}
}

@inproceedings{wang2024mathvision,
  author    = {Ke Wang and Junting Pan and Weikang Shi and Zimu Lu and Houxing Ren and Aojun Zhou and Mingjie Zhan and Hongsheng Li},
  title     = {Measuring Multimodal Mathematical Reasoning with {MATH-Vision} Dataset},
  booktitle = {Advances in Neural Information Processing Systems},
  volume    = {37},
  pages     = {95095--95169},
  year      = {2024},
  doi       = {10.52202/079017-3014},
  url       = {https://proceedings.neurips.cc/paper_files/paper/2024/hash/ad0edc7d5fa1a783f063646968b7315b-Abstract-Datasets_and_Benchmarks_Track.html}
}

@inproceedings{yue2025mmmupro,
  author    = {Xiang Yue and Tianyu Zheng and Yuansheng Ni and Yubo Wang and Kai Zhang and Shengbang Tong and Yuxuan Sun and Botao Yu and Ge Zhang and Huan Sun and Yu Su and Wenhu Chen and Graham Neubig},
  title     = {{MMMU}-Pro: A More Robust Multi-discipline Multimodal Understanding Benchmark},
  booktitle = {Proceedings of the 63rd Annual Meeting of the Association for Computational Linguistics (Volume 1: Long Papers)},
  pages     = {15134--15186},
  year      = {2025},
  publisher = {Association for Computational Linguistics},
  doi       = {10.18653/v1/2025.acl-long.736},
  url       = {https://aclanthology.org/2025.acl-long.736/}
}

@inproceedings{he2025selfcorrection,
  author    = {Jiayi He and Hehai Lin and Qingyun Wang and Yi R. Fung and Heng Ji},
  title     = {Self-Correction is More than Refinement: A Learning Framework for Visual and Language Reasoning Tasks},
  booktitle = {Findings of the Association for Computational Linguistics: ACL 2025},
  pages     = {6405--6421},
  year      = {2025},
  publisher = {Association for Computational Linguistics},
  doi       = {10.18653/v1/2025.findings-acl.331},
  url       = {https://aclanthology.org/2025.findings-acl.331/}
}

@inproceedings{guan2024hallusionbench,
  author    = {Tianrui Guan and Fuxiao Liu and Xiyang Wu and Ruiqi Xian and Zongxia Li and Xiaoyu Liu and Xijun Wang and Lichang Chen and Furong Huang and Yaser Yacoob and Dinesh Manocha and Tianyi Zhou},
  title     = {{HallusionBench}: An Advanced Diagnostic Suite for Entangled Language Hallucination and Visual Illusion in Large Vision-Language Models},
  booktitle = {Proceedings of the IEEE/CVF Conference on Computer Vision and Pattern Recognition},
  pages     = {14375--14385},
  year      = {2024},
  url       = {https://openaccess.thecvf.com/content/CVPR2024/html/Guan_HallusionBench_An_Advanced_Diagnostic_Suite_for_Entangled_Language_Hallucination_and_CVPR_2024_paper.html}
}

@inproceedings{leng2024vcd,
  author    = {Sicong Leng and Hang Zhang and Guanzheng Chen and Xin Li and Shijian Lu and Chunyan Miao and Lidong Bing},
  title     = {Mitigating Object Hallucinations in Large Vision-Language Models through Visual Contrastive Decoding},
  booktitle = {Proceedings of the IEEE/CVF Conference on Computer Vision and Pattern Recognition},
  pages     = {13872--13882},
  year      = {2024},
  url       = {https://openaccess.thecvf.com/content/CVPR2024/html/Leng_Mitigating_Object_Hallucinations_in_Large_Vision-Language_Models_through_Visual_Contrastive_CVPR_2024_paper.html}
}

@inproceedings{vaswani2017attention,
  author       = {Ashish Vaswani and
                  Noam Shazeer and
                  Niki Parmar and
                  Jakob Uszkoreit and
                  Llion Jones and
                  Aidan N. Gomez and
                  Lukasz Kaiser and
                  Illia Polosukhin},
  editor       = {Isabelle Guyon and
                  Ulrike von Luxburg and
                  Samy Bengio and
                  Hanna M. Wallach and
                  Rob Fergus and
                  S. V. N. Vishwanathan and
                  Roman Garnett},
  title        = {Attention is All you Need},
  booktitle    = {Advances in Neural Information Processing Systems 30: Annual Conference
                  on Neural Information Processing Systems 2017, December 4-9, 2017,
                  Long Beach, CA, {USA}},
  pages        = {5998--6008},
  year         = {2017},
  url          = {https://proceedings.neurips.cc/paper/2017/hash/3f5ee243547dee91fbd053c1c4a845aa-Abstract.html},
  bibsource    = {dblp computer science bibliography, https://dblp.org}
}

@inproceedings{ji2026causallens,
  author    = {Junyang Ji and Qifan Liu and Wenming Yang and Zhihai He},
  title     = {{CausalLens}: Sensitivity-Guided Multi-Head Causal Intervention for Hallucination Mitigation in Large Vision-Language Models},
  booktitle = {Proceedings of the IEEE/CVF Conference on Computer Vision and Pattern Recognition},
  pages     = {4199--4209},
  year      = {2026},
  url       = {https://openaccess.thecvf.com/content/CVPR2026/html/Ji_CausalLens_Sensitivity-Guided_Multi-Head_Causal_Intervention_for_Hallucination_Mitigation_in_Large_CVPR_2026_paper.html}
}

@inproceedings{zhang2026seeit,
  author    = {Yongchang Zhang and Oliver Ma and Tianyi Liu and Guangquan Zhou and Yang Chen},
  title     = {See It, Say It, Sorted: An Iterative Training-Free Framework for Visually-Grounded Multimodal Reasoning in {LVLM}s},
  booktitle = {Proceedings of the IEEE/CVF Conference on Computer Vision and Pattern Recognition},
  pages     = {11933--11942},
  year      = {2026},
  url       = {https://openaccess.thecvf.com/content/CVPR2026/html/Zhang_See_It_Say_It_Sorted_An_Iterative_Training-Free_Framework_for_CVPR_2026_paper.html}
}

\clearpage
\appendix
\setcounter{secnumdepth}{2}

\noindent\textbf{Appendix Overview.}
\begin{itemize}
    \item Appendix A documents the evaluated checkpoints, inference
    backends, semantic judge, answer-preference score, paired
    estimands, and a compact claim--evidence index.
    \item Appendix B specifies the prompt and conversation construction
    for direct inference, prior-conditioned reflection, textual
    interventions, continuation probes, and FSAF.
    \item Appendix C follows the diagnosis in the main paper: it reports
    the full prior-CoT influence results, analyzes model-variant and
    scale effects, specifies the operational evidence-span
    decomposition, tests sensitivity to prior-CoT length, and provides
    additional content analyses and a cross-table protocol map.
    \item Appendix D details the evidence-preserving continuation
    protocols and the neutral-history control used to distinguish stale
    evidence reuse from generic extra-turn effects.
    \item Appendix E tests robustness to reflection-prompt wording,
    prefix semantics, and naturally occurring model errors.
    \item Appendix F gives the exact FSAF execution procedure,
    implementation scope, design ablations, state-conditioned
    transitions, and evaluation audit.
    \item Appendix G summarizes the construction, composition, and
    quality controls of VS-Bench, and clarifies how its image-pair
    notation maps to our controlled framework.
    \item Appendix H reports cohort construction, per-model eligible
    counts, and coverage audits.
    \item Appendix I provides representative case studies for the
    diagnostic interventions and evidence-preserving continuations.
\end{itemize}

\paragraph{AI Use Disclosure.}
Generative AI tools were used to assist with language editing, code
debugging, and figure drafting. The authors reviewed and verified all
AI-assisted outputs and take full responsibility for the content of
this manuscript.

\section{Implementation and Evaluation Setup}
\label{app:implementation}

\subsection{Evaluated Checkpoints}
\label{app:model-checkpoints}

Tab.~\ref{tab:model-checkpoints} gives the exact public checkpoint
identifiers used in the 16-model diagnosis. The \emph{Type} column
describes the evaluated checkpoint or decoding mode, rather than
inferring a training recipe from the model name. Qwen3.5 and Qwen3.6
provide unified post-trained checkpoints; we evaluated their
non-thinking mode. For Gemma 4, the suffix \texttt{-it} denotes the
instruction-tuned checkpoint, while the checkpoint without that suffix
is the base variant.

\begin{table*}[t]
\centering
\scriptsize
\setlength{\tabcolsep}{4pt}
\renewcommand{\arraystretch}{1.08}
\begin{tabular}{@{}p{0.32\textwidth}p{0.13\textwidth}p{0.4\textwidth}@{}}
\toprule
\textbf{Checkpoint} & \textbf{Type} & \textbf{Hugging Face repository} \\
\midrule
\texttt{Qwen/Qwen2.5-VL-7B-Instruct}
& Instruct & \url{https://huggingface.co/Qwen/Qwen2.5-VL-7B-Instruct} \\
\texttt{Qwen/Qwen3-VL-8B-Instruct}
& Instruct & \url{https://huggingface.co/Qwen/Qwen3-VL-8B-Instruct} \\
\texttt{Qwen/Qwen3-VL-8B-Thinking}
& Thinking & \url{https://huggingface.co/Qwen/Qwen3-VL-8B-Thinking} \\
\texttt{Qwen/Qwen3-VL-32B-Instruct}
& Instruct & \url{https://huggingface.co/Qwen/Qwen3-VL-32B-Instruct} \\
\texttt{Qwen/Qwen3-VL-32B-Thinking}
& Thinking & \url{https://huggingface.co/Qwen/Qwen3-VL-32B-Thinking} \\
\texttt{Qwen/Qwen3.5-4B}
& Unified (NT) & \url{https://huggingface.co/Qwen/Qwen3.5-4B} \\
\texttt{Qwen/Qwen3.5-9B}
& Unified (NT) & \url{https://huggingface.co/Qwen/Qwen3.5-9B} \\
\texttt{Qwen/Qwen3.5-27B}
& Unified (NT) & \url{https://huggingface.co/Qwen/Qwen3.5-27B} \\
\texttt{Qwen/Qwen3.6-27B}
& Unified (NT) & \url{https://huggingface.co/Qwen/Qwen3.6-27B} \\
\texttt{OpenGVLab/InternVL2-8B}
& Instruct & \url{https://huggingface.co/OpenGVLab/InternVL2-8B} \\
\texttt{google/gemma-4-12B}
& Base & \url{https://huggingface.co/google/gemma-4-12B} \\
\texttt{google/gemma-4-12B-it}
& Instruct & \url{https://huggingface.co/google/gemma-4-12B-it} \\
\texttt{google/gemma-4-31B}
& Base & \url{https://huggingface.co/google/gemma-4-31B} \\
\texttt{google/gemma-4-31B-it}
& Instruct & \url{https://huggingface.co/google/gemma-4-31B-it} \\
\texttt{moonshotai/Kimi-VL-A3B-Instruct}
& Instruct & \url{https://huggingface.co/moonshotai/Kimi-VL-A3B-Instruct} \\
\path{moonshotai/Kimi-VL-A3B-Thinking-2506}
& Thinking & \url{https://huggingface.co/moonshotai/Kimi-VL-A3B-Thinking-2506} \\
\bottomrule
\end{tabular}
\caption{\textbf{Exact model checkpoints used in the 16-model
diagnosis.} Each URL resolves to the corresponding public Hugging Face
repository. NT denotes non-thinking decoding for a unified
post-trained checkpoint.}
\label{tab:model-checkpoints}
\end{table*}

\subsection{Inference Backends and Hardware}
\label{app:inference-backends}

We use vLLM for standard Qwen2.5-VL/Qwen3-VL inference and
Transformers for the remaining model families and probes that require
direct control of rendered prefixes or candidate logits. The semantic
judge is served through an OpenAI-compatible vLLM endpoint, while FSAF
uses Transformers eager-attention hooks
(Appendix~\ref{app:fsaf-implementation}).

Model inference and probing were run on a server with two Intel Xeon
6767P CPUs, 800 GiB of system memory, and eight NVIDIA B200 GPUs
(\(180\) GB each), under CentOS Linux 7 with Linux kernel 6.6.0. The
235B semantic judge was served on four B200 GPUs with tensor parallelism
of \(4\). The software environment used CUDA 12.8, PyTorch 2.10.0,
vLLM 0.19.0, and model-family-specific Transformers environments
ranging from version 4.49.0 to 5.13.0. Within every paired comparison,
the model, image preprocessing, decoding configuration, and maximum
generation length are held fixed; only the specified intervention
changes.

\subsection{Generation and Chat-Template Controls}
\label{app:generation-controls}

Each checkpoint is rendered with its own processor and native chat
template. Images occupy the same conversation position in matched
conditions, and the paired images have identical resolution
(Appendix~\ref{app:vsbench}). We do not rewrite model-specific special
tokens into a shared literal string. Instead, Appendix~\ref{app:prompt-construction}
uses abstract role markers to expose the common conversation structure.
For stochastic decoding runs, all conditions within a matched
model--sample comparison share the same decoding hyperparameters.
FSAF and its ordinary-reflection baseline use a maximum of 256 generated
tokens; the final semantic judge uses temperature zero and a strict
JSON schema.

\subsection{Recorded Provenance and Determinism}
\label{app:recorded-provenance}

Every result row is keyed by model, source dataset, and sample ID.
Intervention outputs additionally record the condition name and the
targets \(y\) and \(y^{-}\). Coverage audits check expected task IDs,
duplicates, missing records, finite scores, and complete matched
condition sets. When a manipulation is randomized, the seed is derived
from the sample identity and is shared by the relevant matched
conditions. Aggregate rates are always recomputed from the retained
row-level records rather than copied from console logs.

\subsection{Semantic-Judge Protocol}
\label{app:semantic-judge}

We use \texttt{Qwen3-VL-235B-A22B-Instruct} as a text-only semantic
judge. The judge receives the dataset name, target answer or target
pair, and the evaluated generation; it does not receive an image.
Its instruction accepts semantically equivalent strings, equivalent
numeric forms (including percentages and decimals when unambiguous),
and multiple-choice letters that identify the same option. For a
two-target task, it independently returns Boolean
\texttt{matches\_new} and \texttt{matches\_old}; for a single-target
task, it returns \texttt{correct}. Requests use batch size one,
temperature zero, and a strict JSON schema that forbids rationales,
extracted text, task IDs, or extra fields. Invalid responses are
retried and are not silently converted to negative labels.

Because the two decisions are independent, VUR and PAR are not forced
to sum to one. Rule-based matching is retained only for inexpensive
design screens explicitly identified as such.

\subsection{Rare Dual-Match Outcomes}
\label{app:dual-match}

Independent target judgments allow a response to match both \(y\) and
\(y^{-}\), for example when it mentions or contrasts both candidates.
We label this outcome \emph{Both}. It contributes to both VUR and PAR
and is therefore retained in all aggregate statistics. However, it is
rare in the reported free-generation analyses: its model-equal share is
\(0.60\%\)--\(1.47\%\) across the evidence-content conditions in
Tab.~\ref{tab:evidence-freegen-distribution}, and only \(10/1{,}800\)
ordinary-reflection outputs (\(0.56\%\)) fall into this category in the
FSAF evaluation (Tab.~\ref{tab:fsaf-state-transitions}). We therefore
do not interpret \emph{Both} as a separate outcome state in the main
text, while retaining its complete counts in the appendix.

\subsection{Answer-Preference Scoring Details}
\label{app:answer-preference}

Answer preference is measured by teacher forcing the canonical current
and prior candidates under the same rendered prefix. When both targets
have valid multiple-choice letters, both candidates are represented by
letters; otherwise both use answer values. Thus, a pair never compares
an option letter on one side with a long answer string on the other.
We score a candidate sequence by the mean log-probability of its
candidate tokens,
\[
  s(a\mid c)=\frac{1}{|a|}\sum_{t=1}^{|a|}
  \log p(a_t\mid c,a_{<t}),
\]
and report \(m(c)=s(y\mid c)-s(y^{-}\mid c)\). Mean rather than summed
log-probability reduces the mechanical length advantage of shorter
candidates. The two-way normalized readout
\[
  \widetilde p(y\mid c)
  =\frac{\exp s(y\mid c)}
  {\exp s(y\mid c)+\exp s(y^{-}\mid c)}
\]
is a relative preference between the two fixed candidates, not a
calibrated probability over all possible answers.

For the FSAF evaluation, every condition appends the candidate-neutral
scaffold \texttt{Final answer:}. If an earlier thinking trace leaves an
open \texttt{<think>} block, it is closed before appending the shared
scaffold. Candidate strings, tokenization rules, and scoring positions
are identical across ordinary reflection, re-query-only, and
FSAF.

\subsection{Why Paired Effects Are Primary}
\label{app:paired-effects}

Our primary estimand is the within-model, within-sample change between
matched conditions. Models differ sharply in baseline calibration,
output style, and eligible-cohort composition, so a raw average across
models can mix the intervention effect with between-model differences.
Pairing removes each model--sample baseline and asks the narrower
question required by the experiments: what changes when only the
history, evidence span, continuation, or attention access changes?
Cross-model averages are therefore descriptive summaries of the paired
effects, not substitutes for them.

\subsection{Claim--Evidence Index}
\label{app:claim-evidence-index}

Tab.~\ref{tab:claim-evidence-index} provides a compact lookup from each
scientific question to its matched comparison, readout, and detailed
protocol.

\begin{table*}[t]
\centering
\scriptsize
\setlength{\tabcolsep}{4pt}
\renewcommand{\arraystretch}{1.10}
\begin{tabular}{@{}p{0.18\textwidth}p{0.22\textwidth}p{0.25\textwidth}p{0.25\textwidth}@{}}
\toprule
\textbf{Question} & \textbf{Primary comparison} &
\textbf{Primary readout} & \textbf{Detailed location} \\
\midrule
Does prior CoT affect reflection?
& Prior-conditioned vs.\ direct/current controls
& VUR, PAR, answer-preference margin
& Appendices~\ref{app:prior-influence} and \ref{app:prompt-carriers} \\
Which prior content carries the effect?
& Intact vs.\ evidence-deleted, answer-only, and matched controls
& Paired change in semantic outcomes and margin
& Appendix~\ref{app:evidence-span-validity} and Tab.~\ref{tab:evidence-organization} \\
Does stale control persist after recovery?
& History-conditioned vs.\ direct-new and neutral-history continuations
& Current/prior preference and semantic branch outcomes
& Appendices~\ref{app:continuation-protocols}--\ref{app:neutral-history} \\
Can current support suppress a stale state?
& Matched support present vs.\ withdrawn
& Prior-answer resurgence within the recovered cohort
& Tab.~\ref{tab:shortcut-retirement} and Appendix~\ref{app:prompt-construction} \\
Does stale visual content transfer?
& History-conditioned vs.\ direct-new/neutral-history derived questions
& Derived-answer preference and semantic accuracy
& Appendix~\ref{app:derived-computation} \\
Does FSAF block prior-CoT access and improve immediate updating?
& FSAF vs.\ ordinary reflection on identical model--sample pairs
& Paired immediate VUR, PAR, and answer preference
& Appendices~\ref{app:fsaf-implementation}--\ref{app:fsaf-audit} \\
Which rows enter each analysis?
& Directly solvable pair \(\rightarrow\) experiment-specific filters
& Eligible and complete-pair counts
& Appendix~\ref{app:cohort-coverage} \\
\bottomrule
\end{tabular}
\caption{\textbf{Index from the paper's main claims to protocols,
metrics, and appendix details.}}
\label{tab:claim-evidence-index}
\end{table*}

\section{Prompt and Conversation Construction}
\label{app:prompt-construction}

The notation below follows the abstract, model-independent rendering
used to compare different chat templates. \texttt{[User Start]},
\texttt{[User End]}, \texttt{[Response Start]}, and
\texttt{[Response End]} denote role boundaries inserted by each
model's native processor; they are not literal strings passed to every
checkpoint. \(I^{-}\) and \(I\) are the prior and current images,
\(R^{-}\) is the prior response, \(R\) is the reflected response, and
\(Q'\) is a controlled continuation question.

\subsection{Direct Inference}

\noindent\textbf{(1) Prior-image direct inference.}
The model receives \(I^{-}\) and \(Q\), producing \(R^{-}\) in one
assistant turn.
\begin{itemize}
    \item \textbf{Input:}
    \begin{center}\small\ttfamily
    [User Start] \(I^{-}\ Q\) [User End]
    \end{center}
    \item \textbf{Output:}
    \begin{center}\small\ttfamily
    [Response Start] \(R^{-}\) [Response End]
    \end{center}
\end{itemize}

\noindent\textbf{(2) Current-image direct inference.}
The same construction replaces \(I^{-}\) with \(I\), yielding the
direct-current response \(R_{\mathrm{dir}}\).
\begin{itemize}
    \item \textbf{Input:}
    \begin{center}\small\ttfamily
    [User Start] \(I\ Q\) [User End]
    \end{center}
    \item \textbf{Output:}
    \begin{center}\small\ttfamily
    [Response Start] \(R_{\mathrm{dir}}\) [Response End]
    \end{center}
\end{itemize}
These two runs define eligibility; they are not counted as reflection
successes.

\subsection{Prior-Conditioned Reflection}

\noindent\textbf{(3) Standard reflection.}
We retain the current image and question, prefill the assistant with the
complete prior response followed by the fixed cue
\begin{quote}\small
\texttt{Wait, let me check the figure again to make sure I haven't made a mistake.}
\end{quote}
and ask the model to continue the same assistant turn.
\begin{itemize}
    \item \textbf{Input:}
    \begin{center}\small\ttfamily
    [User Start] \(I\ Q\) [User End]\\{}
    [Response Start] \(R^{-}\) [Reflection Cue]
    \end{center}
    \item \textbf{Output:}
    \begin{center}\small\ttfamily
    [Response Start] \(R^{-}\) [Reflection Cue] \(R\) [Response End]
    \end{center}
\end{itemize}
The prior text is therefore part of the cached assistant prefix, while
the image changes from \(I^{-}\) to \(I\).

\subsection{Prior-Carrier and Evidence-Content Interventions}
\label{app:prompt-carriers}

\noindent\textbf{(4) Carrier ablations.}
All conditions retain \(I\), \(Q\), and the same reflection cue. Only
the assistant prefix \(C\) changes:
\begin{quote}\small
\texttt{full prior:} \(C=R^{-}\);\quad
\texttt{reasoning only:} \(C=R^{-}\) without its final answer;\quad
\texttt{answer only:} \texttt{My previous answer was }\(y^{-}\);\quad
\texttt{answer masked:} the reasoning plus
\texttt{My initial answer is [omitted].}
\end{quote}
The random-answer control replaces \(y^{-}\) with a non-target answer,
and the new-turn control preserves \(R^{-}\) as a completed historical
assistant message rather than a same-turn prefill. Their shared schema
is:
\begin{center}\small\ttfamily
[User Start] \(I\ Q\) [User End]\\{}
[Response Start] \(C\) [Reflection Cue] \(R\) [Response End].
\end{center}

\noindent\textbf{(5) Evidence-content interventions.}
Sentence-level interventions replace \(C\) with an intact, deleted,
masked, substituted, or organization-controlled version of the prior
reasoning. The image, question, cue, targets, decoding settings, and
sample identity remain fixed. Evidence-deletion controls are matched
either by removed-sentence count or approximate removed-token length;
generic-memory controls preserve a prior-context carrier without
retaining answer-relevant visual content. Exact evidence-span rules are
given in Appendix~\ref{app:evidence-span-validity}.

\subsection{Matched Support-Withdrawal Test}

\noindent\textbf{(6) Current support present vs.\ withdrawn.}
Within samples whose ordinary reflection reaches the current answer,
the support-present condition retains the complete current-evidence
state. The matched withdrawal condition deletes all current
evidence-bearing spans while retaining the model, sample, current image,
question, stale history, answer scaffold, and target candidates. We
apply two complementary readouts to every matched context:
teacher-forced candidate scoring measures answer preference, while a
separate continuation is evaluated by the semantic judge to obtain the
free-generation PAR reported in Tab.~\ref{tab:shortcut-retirement}.
The D/H paths and F/W conditions retain the same model--sample identity,
so the generated outcomes are matched rather than unrelated samples.
The \emph{recovered} label is assigned from the source ordinary-reflection
output before the F/W contexts are reconstructed. Tab.~\ref{tab:shortcut-retirement}
then evaluates newly generated continuations from those contexts; hence
the full-support H continuation can match \(y^{-}\) even though its source
state previously produced \(y\). Because VUR and PAR are judged
independently, PAR also includes any continuation matching both targets.
This construction isolates whether the recovered state remains stable
when its current textual support is removed; it does not remove the
current image.

\subsection{Evidence-Preserving Continuations}

\noindent\textbf{(7) Future re-query.}
After constructing a source state, we close that turn and append one of
two user questions:
\begin{quote}\small
\texttt{Equivalent re-query:} ``For the current image, answer the same
question again. Use the current image as the visual evidence.''\\
\texttt{Concise re-query:} ``For the current image, give only the final
answer to the question below.''
\end{quote}
The original question \(Q\) follows either instruction. The resulting
conversation is:
\begin{center}\small\ttfamily
[Source-State Conversation]\\{}
[User Start] \(I\ Q'\) [User End]\\{}
[Response Start] \(R'\) [Response End].
\end{center}
For derived questions, \(Q'\) requests a new computation from an
answer-critical visual premise rather than simply asking for \(y\)
again. The source-state construction and target derivation are held
fixed across direct-new, history-conditioned, and neutral-history
conditions.

\subsection{Neutral-History Control}

\noindent\textbf{(8) Content-neutral history.}
The control preserves a multi-turn history but removes its concrete
visual premises. Its first user message states:
\begin{quote}\small
\texttt{Earlier there was a related context, but it should not provide
visual evidence for the current image.}
\end{quote}
The historical assistant content is replaced by the repeated neutral
sentence
\begin{quote}\small
\texttt{I considered an earlier context, but its visual details are not
evidence for the current image.}
\end{quote}
up to the capped number of prior sentences. The conversation then
contains the direct-current \(I,Q\) exchange before \(Q'\) is appended.
This retains extra roles and history structure without retaining the
specific stale evidence-bearing content.

\subsection{FSAF Conversation}

\noindent\textbf{(9) Fresh-state re-query with attention firewall.}
Ordinary reflection first produces the stale assistant prefix. FSAF
then creates a fresh user turn:
\begin{quote}\small
\texttt{Answer the same question using only the current image. Be
concise and end with the final-answer format requested by the question.}
\end{quote}
and a fresh assistant suffix. During both prefill and generation of this
suffix, the firewall blocks attention from fresh-suffix queries to the
identified stale assistant span:
\begin{center}\small\ttfamily
[Stale-State Conversation]\\{}
[User Start] [FSAF Re-query] [User End]\\{}
[Response Start] \(R_{\mathrm{FSAF}}\) [Response End].
\end{center}
The current image remains available through the current conversation;
only access to the stale textual span is masked. Appendix
\ref{app:fsaf-implementation} specifies the layerwise mask and verified
model scope.

\section{Diagnosing Textual Shortcuts}
\label{app:diagnosis-details}

This section complements the main-text diagnosis. We first report the
complete cross-model prior-influence control, analyze model-variant and
scale effects, then document how evidence-bearing spans are identified
and intervened on, and finally provide additional free-generation
results and a cross-table protocol map.

\subsection{Full Prior-CoT Influence Across 16 VLMs}
\label{app:prior-influence}

Tab.~\ref{tab:prior-influence-full} reports the complete
semantic-judge results for the control summarized in
Tab.~\ref{tab:prior-influence-main}. Prior-only PAR is the rate of
generating the prior answer when \(Q\) and \(R^{-}\) are retained but
the current image is withheld. Image-present VUR is the rate of
generating the current answer after restoring \(I\) under the matched
new-turn setting. In both columns, \(R^{-}\) is preserved as a completed
historical assistant response and a fresh user turn re-asks \(Q\);
the two columns differ only in whether that turn includes \(I\). This
new-turn control is distinct from the same-assistant-turn prefix used
for standard reflection earlier in Appendix~B and for the N/C/E
evidence-content interventions in
Tab.~\ref{tab:evidence-freegen-distribution}. Their absolute VUR values
therefore estimate different conversational carriers and are not
expected to coincide. This distinction also accounts for the apparent
discrepancy for Qwen3-VL-8B-Instruct: the standard-reflection VUR in
Tab.~\ref{tab:prompt-prefix-robustness} is \(49.1\%\) averaged over
prompt paraphrases and \(49.5\%\) with the fixed reflection prefix,
whereas the image-present new-turn VUR here is \(44.9\%\). In the former,
the model continues an unfinished assistant prefix under the
current-image context; in the latter, \(R^{-}\) is presented as a
completed historical answer before \(Q\) is asked again. The direction
of this difference is consistent with the Qwen2.5-VL-7B rule-based
FSAF design screen in Appendix~\ref{app:fsaf-ablation}: a fresh
re-query boundary alone does
not isolate new computation from \(R^{-}\), because the re-query tokens
can still absorb stale information during prefill and carry it into
generation through their cached states. In that auxiliary screen,
re-query only
provides no reliable VUR gain while increasing PAR, whereas full FSAF
couples the fresh boundary with a prefill- and generation-time
attention firewall. The lower new-turn VUR is therefore compatible
with role separation alone failing to retire the textual shortcut,
rather than evidence of a repeated-measurement mismatch. This
interpretation is descriptive, not a causal estimate, because the
robustness and full-control summaries also use their respective
experiment-specific eligibility cohorts.
Prior-only PAR is high for all 16 models, whereas image-present VUR
varies substantially, showing that the prior CoT provides strong answer
control even though current visual evidence remains behaviorally
effective.

\begin{table}[h!]
\centering
\scriptsize
\begin{tabular}{llcc}
\toprule
\textbf{Model} & \textbf{Variant}
& \shortstack{\textbf{Prior-only}\\\textbf{PAR (\%)}}
& \shortstack{\textbf{Image-present}\\\textbf{VUR (\%)}} \\
\midrule
Qwen2.5-VL-7B       & Instruct & 74.8 & 48.4 \\
\multirow{2}{*}{Qwen3-VL-8B}
                    & Instruct & 70.8 & 44.9 \\
                    & Thinking & 53.3 & 22.6 \\
\multirow{2}{*}{Qwen3-VL-32B}
                    & Instruct & 82.9 & 60.0 \\
                    & Thinking & 56.6 & 31.8 \\
Qwen3.5-4B          & --       & 98.7 & 82.0 \\
Qwen3.5-9B          & --       & 99.2 & 86.8 \\
Qwen3.5-27B         & --       & 97.8 & 88.0 \\
Qwen3.6-27B         & --       & 97.8 & 85.5 \\
InternVL2-8B        & Instruct & 87.9 & 38.3 \\
\multirow{2}{*}{Gemma-4-12B}
                    & Base     & 95.0 & 75.0 \\
                    & IT       & 91.3 & 79.9 \\
\multirow{2}{*}{Gemma-4-31B}
                    & Base     & 97.1 & 77.9 \\
                    & IT       & 92.7 & 88.0 \\
\multirow{2}{*}{Kimi-VL-A3B}
                    & Instruct & 97.2 & 56.9 \\
                    & Thinking & 96.7 & 52.5 \\
\bottomrule
\end{tabular}
\caption{\textbf{Prior-CoT influence and visual availability across all
16 VLMs.} Rates are percentages over the eligible cohort for each
model and are evaluated by the semantic judge.}
\label{tab:prior-influence-full}
\end{table}

\subsection{Thinking versus Instruct Models and Scale}
\label{app:variant-scale}

VisualSwap reports that Thinking variants suffer larger base--probe
accuracy degradation than their Instruct counterparts and that
increasing model size does not mitigate this degradation
\citep{shi2026visualswap}. We revisit both observations using our
intervention-based readouts. The comparisons below are descriptive:
variants and scales use their own eligible cohorts, and our VUR/PAR
readouts are not numerically equivalent to VisualSwap's base--probe
accuracy gap.

\begin{table}[h!]
\centering
\scriptsize
\setlength{\tabcolsep}{2.4pt}
\renewcommand{\arraystretch}{1.08}
\resizebox{\columnwidth}{!}{%
\begin{tabular}{@{}llrrrr@{}}
\toprule
\textbf{Model}
& \textbf{Variant}
& \shortstack{\textbf{Prior-only}\\\textbf{PAR (\%)}}
& \shortstack{\textbf{Image-present}\\\textbf{VUR (\%)}}
& \(\Delta m_{\mathrm{E-C}}\)
& \shortstack{\(\Delta\)\textbf{VUR / }\(\Delta\)\textbf{PAR}\\\textbf{E--C (pp)}} \\
\midrule
\multirow{2}{*}{Qwen3-VL-8B}
& Instruct & 70.8 & 44.9 & +2.617 & \(+42.27/-72.95\) \\
& Thinking & 53.3 & 22.6 & +2.832 & \(+35.39/-84.65\) \\
\midrule
\multirow{2}{*}{Qwen3-VL-32B}
& Instruct & 82.9 & 60.0 & +3.075 & \(+54.74/-81.52\) \\
& Thinking & 56.6 & 31.8 & +3.004 & \(+54.47/-82.11\) \\
\midrule
\multirow{2}{*}{Kimi-VL-A3B}
& Instruct & 97.2 & 56.9 & +1.619 & \(+32.04/-49.17\) \\
& Thinking & 96.7 & 52.5 & +2.913 & \(+51.91/-66.12\) \\
\bottomrule
\end{tabular}
}
\caption{\textbf{Matched-family comparison of Instruct and Thinking
variants.}
Prior-only PAR and image-present VUR are absolute semantic-judge rates.
E--C removes evidence-bearing spans rather than length-matched
non-evidence context; positive margin/VUR and negative PAR changes
indicate weaker prior-answer control.}
\label{tab:variant-comparison}
\end{table}

\noindent\textbf{Thinking versus Instruct.}
Thinking yields lower image-present VUR in all three same-family,
same-checkpoint comparisons: by \(22.3\) pp at Qwen3-VL-8B,
\(28.2\) pp at Qwen3-VL-32B, and \(4.4\) pp for Kimi-VL-A3B.
The pair-equal average is \(35.6\%\) for Thinking versus \(53.9\%\)
for Instruct, agreeing directionally with VisualSwap's finding that
Thinking variants are more vulnerable during continuous
re-examination. However, Thinking does not exhibit higher prior-only
PAR: it is lower in all three pairs. The mode gap is therefore more
specific than a universally stronger standalone preference for the
prior answer; it appears when the current image must compete with the
retained reasoning trajectory.

The intervention results refine this comparison. Evidence removal
shifts \(m\) toward the current answer for every variant, with
pair-equal \(\Delta m_{\mathrm{E-C}}\) of \(2.916\) for Thinking and
\(2.437\) for Instruct. It also produces a larger PAR reduction for
Thinking in all three pairs, whereas the corresponding VUR gain is
mixed across families. Thus, evidence-bearing prior content is the
shared intervention carrier in both modes, but releasing its
prior-answer control does not always translate into correct recovery.

\begin{table}[h!]
\centering
\scriptsize
\setlength{\tabcolsep}{2.0pt}
\renewcommand{\arraystretch}{1.08}
\begin{tabular*}{\columnwidth}{@{\extracolsep{\fill}}lrrrr@{}}
\toprule
\textbf{Family / Variant}
& \textbf{Size}
& \shortstack{\textbf{Prior-only}\\\textbf{PAR (\%)}}
& \shortstack{\textbf{Image-present}\\\textbf{VUR (\%)}}
& \(\Delta m_{\mathrm{E-C}}\) \\
\midrule
\multirow{2}{*}{Qwen3-VL / Instruct}
& 8B  & 70.8 & 44.9 & 2.617 \\
& 32B & 82.9 & 60.0 & 3.075 \\
\midrule
\multirow{2}{*}{Qwen3-VL / Thinking}
& 8B  & 53.3 & 22.6 & 2.832 \\
& 32B & 56.6 & 31.8 & 3.004 \\
\midrule
\multirow{3}{*}{Qwen3.5}
& 4B  & 98.7 & 82.0 & 1.612 \\
& 9B  & 99.2 & 86.8 & 1.372 \\
& 27B & 97.8 & 88.0 & 1.535 \\
\midrule
\multirow{2}{*}{Gemma-4 / Base}
& 12B & 95.0 & 75.0 & 0.911 \\
& 31B & 97.1 & 77.9 & 0.996 \\
\midrule
\multirow{2}{*}{Gemma-4 / IT}
& 12B & 91.3 & 79.9 & 2.067 \\
& 31B & 92.7 & 88.0 & 3.152 \\
\bottomrule
\end{tabular*}
\caption{\textbf{Descriptive within-family scale trends.}
Rows within each block follow increasing nominal checkpoint size.
Cohorts are separately eligible at each checkpoint, so these trends
are not paired sample-level scaling estimates.}
\label{tab:scale-comparison}
\end{table}

\noindent\textbf{Scale.}
Image-present VUR increases with nominal size in all five comparable
within-family sequences in Tab.~\ref{tab:scale-comparison}. This
differs from a literal replication of VisualSwap's degradation trend,
but it does not show shortcut retirement. Prior-only PAR increases in
four of the five sequences and remains near saturation for Qwen3.5,
while \(\Delta m_{\mathrm{E-C}}\) does not decrease monotonically with
size. Larger checkpoints can therefore recover the current answer more
often while retaining an equally strong or stronger evidence-bearing
prior path. In particular, Qwen3-VL-32B-Thinking remains far below its
Instruct counterpart in VUR (\(31.8\%\) versus \(60.0\%\)).

FSAF further separates vulnerability from immutable visual ability.
On the independently eligible FSAF cohort in
Tab.~\ref{tab:fsaf-main}, it raises VUR more for Thinking than
Instruct at both Qwen3-VL scales
(\(+19.96\) versus \(+15.95\) pp at 8B and \(+21.97\) versus
\(+15.22\) pp at 32B), while reducing PAR to \(1.03\%\)--\(1.64\%\)
for the Thinking variants. The remaining post-FSAF VUR gap shows that
blocking prior-CoT access does not erase every mode difference, but the
large matched improvement identifies stale textual accessibility as a
substantial and scale-persistent component of the Thinking-model
failure.

\subsection{Operational Identification of Evidence-Bearing Spans}
\label{app:evidence-span-validity}

We identify \(E^{-}\) automatically using fixed, outcome-independent
rules. Before classification, we remove the explicit final-answer
suffix using deterministic patterns for expressions such as
``final answer,'' ``the answer is,'' and
\texttt{\textbackslash boxed};
when none matches and the response contains more than two sentences,
the implementation removes the final sentence. This preprocessing keeps
the explicit answer span \(A^{-}\) separate from the reasoning spans
considered for \(E^{-}\). We then split the remaining prior CoT at
sentence-final periods, question marks, exclamation marks, and line
breaks.

A sentence is marked as evidence-bearing if it satisfies at least one
of the following four rules:
\begin{enumerate}
    \item it contains a number, \texttt{\textbackslash frac},
    \texttt{\textbackslash sqrt}, or one of
    \(=, <, >, \pm,\) and degree symbols;
    \item it contains an explicit option or answer reference involving
    a letter from A to J;
    \item it contains a term from the fixed visual lexicon:
    \emph{image, figure, graph, diagram, line, point, circle, triangle,
    angle, axis, x-axis, y-axis, table, chart, shape, red, blue, green,
    left, right, above, below, intersect, parallel, perpendicular,
    slope, curve, bar, row, column, object, shown, visible}; or
    \item it contains a term from the fixed mathematical/reasoning
    lexicon: \emph{equals, equal, therefore, thus, hence, because,
    since, so, calculate, compute, sum, difference, ratio, area,
    perimeter, length, radius, diameter, probability, fraction, degree,
    option, choice, answer, value}.
\end{enumerate}
For the dose-response intervention in
Fig.~\ref{fig:evidence-dose-response}, identified sentences are ranked
by the number of numerical or symbolic cues, twice the number of option
cues, and the number of matched visual and mathematical terms; ties are
resolved by sentence length. At removal fraction \(\alpha\), we delete
the highest-ranked \(\lceil\alpha |E^{-}|\rceil\) sentences. The
length-matched control randomly orders non-evidence sentences using a
fixed sample seed and removes sentences until their token count reaches
that of the removed evidence. These rules deliberately provide a
reproducible, high-recall operational decomposition; they do not claim
that the selected sentences are unique human-annotated causal spans.

\subsection{Free-Generation Outcomes after Evidence Removal}
\label{app:evidence-free-generation}

Tab.~\ref{tab:evidence-freegen-distribution} reports the complete
model-equal outcome distribution underlying the free-generation
results in Tab.~\ref{tab:evidence-content}. The intact prior (N)
retains \(R^{-}\), the matched context control (C) deletes a
length-matched amount of non-evidence context, and evidence removal (E)
deletes all spans in \(E^{-}\). All three are generated by continuing
the same assistant turn after the edited prefix and fixed reflection
cue. Thus, intact N is not the image-present new-turn condition in
Tab.~\ref{tab:prior-influence-full}; the appropriate evidence-content
comparison is the matched E--C contrast reported in
Tab.~\ref{tab:evidence-content}, not a cross-table comparison of their
absolute VUR values.

\begin{table}[h!]
\centering
\scriptsize
\setlength{\tabcolsep}{3pt}
\begin{tabular}{@{}lcccc@{}}
\toprule
\textbf{Condition}
& \textbf{VUR (\%)}
& \textbf{PAR (\%)}
& \textbf{Both (\%)}
& \textbf{Other (\%)} \\
\midrule
Intact prior (N)             & 14.32 & 78.77 & 1.44 & 8.35 \\
Matched context control (C)  & 15.78 & 77.14 & 1.47 & 8.55 \\
Evidence removal (E)         & 59.54 & 14.55 & 0.60 & 26.50 \\
\bottomrule
\end{tabular}
\caption{\textbf{Free-generation outcome distribution under
evidence-content interventions.}
Entries are model-equal percentages over \(4{,}159\) paired
model--sample instances from 16 VLMs. VUR and PAR are judged
independently; \emph{Both} denotes outputs matching both targets, and
\emph{Other} denotes outputs matching neither target.}
\label{tab:evidence-freegen-distribution}
\end{table}

Relative to the intact prior, evidence removal increases VUR by
\(45.22\) pp and decreases PAR by \(64.22\) pp. The released
prior-answer control does not always become the current answer:
\emph{Other} increases from \(8.35\%\) to \(26.50\%\), a change of
\(18.16\) pp, with a positive change in \(14/16\) models. Thus,
removing \(E^{-}\) suppresses prior-consistent generation but does not
guarantee recovery of the current answer.

\begin{figure}[t]
\centering
\includegraphics[width=\linewidth]{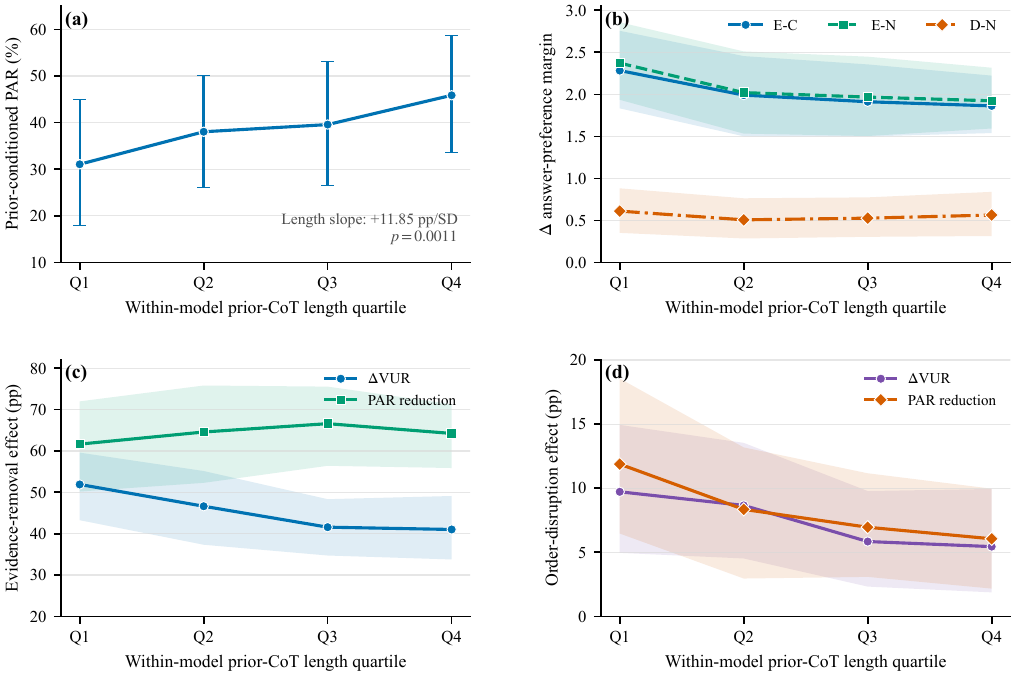}
\caption{\textbf{Prior-CoT length predicts vulnerability but does not
explain the core intervention effects.}
(a) Model-equal PAR under image-present prior-conditioned reflection.
(b) Answer-preference effects for evidence removal versus a
length-matched non-evidence control (E--C), evidence removal versus the
intact prior (E--N), and evidence-order disruption versus the intact
prior (D--N). (c,d) Corresponding VUR increases and PAR reductions.
Error bars and bands are 95\% hierarchical-bootstrap CIs. 
}
\label{fig:prior-cot-length-sensitivity}
\end{figure}

\noindent\textbf{Prior-CoT Length Sensitivity.}
We test whether prior-CoT length can explain the observed vulnerability.
Length is measured on the prior reasoning span \(R^{-}\) after removing
its explicit final-answer suffix. Within each model, we divide the
\(4{,}159\) eligible model--sample instances into character-length
quartiles, preventing naturally longer traces from being treated as a
between-model effect. Under image-present prior-conditioned reflection,
model-equal PAR increases from \(31.07\%\) in the shortest quartile to
\(45.87\%\) in the longest. After controlling for model and dataset
fixed effects, a one-standard-deviation increase in
\(\log(1+\mathrm{length})\) is associated with an \(11.85\) pp increase
in PAR (\(p=0.0011\)). Longer prior CoTs are therefore associated with
greater prior-answer retention.

This association does not account for the content and organization
effects. Across all four within-model length quartiles, evidence
removal remains stronger than length-matched non-evidence removal and
the intact prior, while evidence-order disruption continues to weaken
prior control (Fig.~\ref{fig:prior-cot-length-sensitivity}). The
corresponding answer-preference, VUR, and PAR effects retain their
reported directions, and none of the seven core intervention readouts
has a significant character-length slope after the fixed-effect
adjustment (all \(p>0.05\)). This analysis is an observational
stratification and adjustment, rather than a same-sample intervention
on prior-CoT length, and therefore does not establish that length has
no causal role.

\subsection{Protocol Map for Reported Results}
\label{app:result-protocol-map}

Tabs.~\ref{tab:prior-influence-main}--\ref{tab:fsaf-state-transitions}
use several deliberately different carriers and turn structures.
Tab.~\ref{tab:result-protocol-map} makes these settings explicit so
that absolute rates are compared only within a matched protocol.

\begin{table*}[t]
\centering
\scriptsize
\setlength{\tabcolsep}{3pt}
\renewcommand{\arraystretch}{1.12}
\begin{tabular}{@{}p{0.11\textwidth}p{0.25\textwidth}p{0.34\textwidth}p{0.23\textwidth}@{}}
\toprule
\textbf{Table(s)} & \textbf{Carrier} & \textbf{Turn structure} &
\textbf{Evaluator} \\
\midrule
\ref{tab:prior-influence-main}, \ref{tab:prior-influence-full}
& Completed historical \(R^{-}\)
& Fresh user turn re-asks \(Q\), with or without \(I\)
& 235B semantic judge \\
\ref{tab:evidence-content}, \ref{tab:evidence-freegen-distribution}
& Intact or edited same-turn \(R^{-}\) prefix (N/C/E/A)
& Same assistant turn after the edited prefix and fixed cue
& Model token log-probabilities and 235B judge \\
\ref{tab:evidence-organization}
& Intact or reordered evidence chain (N/D/NE)
& Same assistant turn after the controlled prefix
& Model token log-probabilities and 235B judge \\
\ref{tab:shortcut-retirement}
& Reconstructed D/H contexts with current support F/W
& Separate continuation from each matched context
& 235B semantic judge \\
\ref{tab:future-state-continuation}, \ref{tab:neutral-history-control}
& Direct, neutral-history, or prior-conditioned source state
& Fresh equivalent-re-query or derived-computation user turn
& Model token log-probabilities and 235B judge \\
\ref{tab:prompt-prefix-robustness}
& Paraphrased/neutral cue or naturally generated erroneous CoT
& Same-turn reflection or matched prompt-only control
& 235B semantic judge \\
\ref{tab:fsaf-main}, \ref{tab:fsaf-state-transitions}
& Ordinary stale prefix or fresh suffix blocked from \(R^{-}\)
& Same-turn baseline vs.\ fresh user--assistant FSAF boundary
& 235B judge and model token log-probabilities \\
\ref{tab:variant-comparison}, \ref{tab:scale-comparison}
& Derived summaries of the preceding carrier conditions
& Mixed: new-turn absolute rates and same-turn E--C intervention
& Inherited from Tabs.~\ref{tab:prior-influence-full} and \ref{tab:evidence-content} \\
\ref{tab:fsaf-ablation}
& RAQ, CRF, re-query-only, or full FSAF
& Condition-specific same-turn or fresh-boundary design screen
& Rule screen; 235B judge where marked \\
\bottomrule
\end{tabular}
\caption{\textbf{Protocol map for reported quantitative results.}
Eligibility rules and cohort counts are documented separately in
Appendix~\ref{app:cohort-coverage} and are not repeated here.
Descriptive/setup Tabs.~\ref{tab:model-checkpoints},
\ref{tab:claim-evidence-index}, \ref{tab:reflection-prompt-variants},
\ref{tab:vsbench-composition}, \ref{tab:sample-flow}, and
\ref{tab:eligible-counts} are omitted.}
\label{tab:result-protocol-map}
\end{table*}

\section{Shortcut Persistence and Continuation Controls}
\label{app:continuation-protocols}

This section details the two evidence-preserving continuation branches
used to test whether shortcut influence survives answer correction or
extends to later computation. Both branches begin from the matched
direct and prior-conditioned paths in
Eqs.~(\ref{eq:direct-state})--(\ref{eq:history-state}) and preserve the
current image and preceding path-specific context.

Tab.~\ref{tab:future-state-continuation} displays the two initial
outcomes central to the state-conditioned analysis rather than an
exhaustive partition. For equivalent re-query, the \(4{,}159\) eligible
pairs comprise \(2{,}456\) recovered (\(59.05\%\)), \(1{,}418\)
active-stale (\(34.09\%\)), and \(285\) Other (\(6.85\%\)) cases. For
derived computation, the \(2{,}384\) numeric-eligible pairs comprise
\(1{,}450\) recovered (\(60.82\%\)), \(770\) active-stale
(\(32.30\%\)), and \(164\) Other (\(6.88\%\)) cases. The analyzed
source-state partitions contain no Both cases. Other is omitted from the
main table because the two named strata provide the interpretable
current-only and prior-only conditioning states; it is not folded into
either reported effect.

\subsection{Equivalent Re-Query}
\label{app:equivalent-requery}

The equivalent re-query is the first gray continuation branch in
Fig.~\ref{fig:framework}.
We append a new user turn that repeats \(Q\). We use two semantically
matched variants.
The first explicitly instructs the model to use the current image as
visual evidence before answering \(Q\) again; the second requests only
a concise final answer to \(Q\). Candidate scoring compares support for
\(y\) and \(y^{-}\) using \(m(S)\) in
Eq.~(\ref{eq:new-old-margin}), while free generation is evaluated with
VUR and PAR in Eq.~(\ref{eq:vur-par}). Neither variant removes preceding
support or inserts \(y^{-}\) into the new prompt. An auxiliary
neutral-history control preserves the extra turn and context structure
without including the evidence-bearing content of \(R^{-}\).

\subsection{Derived Computation}
\label{app:derived-computation}

For tasks in which the current and prior quantities can be parsed as
distinct numerical values, we replace the repeated question with one
of three transformations: doubling the quantity, adding one, or testing
it against a threshold chosen to distinguish the current- and
prior-derived outcomes. The transformed current and prior values serve
as the paired candidates for scoring and as the targets for judging
free generation under the same metric definitions. No support is
removed and \(y^{-}\) is not reintroduced in the follow-up prompt.

\subsection{Neutral-History Control}
\label{app:neutral-history}

To distinguish the effect of stale evidence from generic effects of an
additional dialogue turn or longer context, we introduce a
neutral-history path that preserves the extra turn and context structure
but excludes the evidence-bearing content of \(R^{-}\).
Tab.~\ref{tab:neutral-history-control} reports free-generation match
rates for the direct, neutral-history, and prior-conditioned paths,
stratified by the initial reflection outcome.

\begin{table*}[t]
\centering
\small
\setlength{\tabcolsep}{4.5pt}
\renewcommand{\arraystretch}{1.10}
\begin{tabular}{@{}llrrrrrr@{}}
\toprule
& & \multicolumn{3}{c}{\textbf{Current match (\%)}}&
\multicolumn{3}{c}{\textbf{Prior match (\%)}} \\
\cmidrule(lr){3-5}\cmidrule(lr){6-8}
\textbf{Continuation}
& \textbf{Initial outcome}
& \textbf{Direct}
& \textbf{Neutral}
& \shortstack{\textbf{Prior-cond.}\\\textbf{(H)}}
& \textbf{Direct}
& \textbf{Neutral}
& \shortstack{\textbf{Prior-cond.}\\\textbf{(H)}} \\
\midrule
\multirow{2}{*}{Equivalent re-query}
& Recovered    & 54.80 & 55.56 & 52.67 & 0.57 & 0.75 & 1.55 \\
& Active stale & 56.77 & 57.48 & 16.89 & 0.88 & 1.02 & 35.93 \\
\midrule
\multirow{2}{*}{Derived computation}
& Recovered    & 72.94 & 67.01 & 53.95 & 4.64 & 5.11 & 4.71 \\
& Active stale & 76.02 & 73.07 & 10.82 & 8.83 & 8.57 & 54.11 \\
\bottomrule
\end{tabular}
\caption{\textbf{Neutral-history control for evidence-preserving
continuations.} For derived computation, current and prior matches refer
to the corresponding derived targets.}
\label{tab:neutral-history-control}
\end{table*}

For equivalent re-query, neutral-history rates remain within \(0.8\) pp
of the direct path for current-answer matches and within \(0.2\) pp for
prior-answer matches in both initial-outcome strata. In contrast, among
active-stale outcomes, prior-conditioned history decreases the
current-answer rate from \(56.77\%\) to \(16.89\%\) and increases the
prior-answer rate from \(0.88\%\) to \(35.93\%\). The recurrence is
therefore associated with the stale evidence-bearing history rather
than the presence of an additional turn alone.

For derived computation, neutral history causes a smaller generic
decrease in current-derived matches relative to the direct path
(\(-5.93\) pp for recovered and \(-2.95\) pp for active-stale outcomes),
but leaves prior-derived matches nearly unchanged (\(+0.47\) and
\(-0.26\) pp, respectively). Prior-conditioned history produces a much
larger active-stale shift, decreasing current-derived matches by
\(65.19\) pp and increasing prior-derived matches by \(45.28\) pp.
For recovered outcomes, it decreases current-derived matches by
\(18.99\) pp while changing prior-derived matches by only \(+0.07\) pp.
Thus, generic history can contribute modestly to computation
degradation, but it does not explain the strong premise transfer in the
active-stale stratum; after recovery, the effect is better characterized
as computation disruption than stable prior-premise substitution.

\section{Robustness Checks and Naturally Occurring Errors}
\label{app:prompt-robustness}

\subsection{Reflection-Prompt and Prefix Robustness}
\label{app:prompt-prefix-robustness}

To test whether prior-CoT influence is induced by a particular
reflection phrase, we follow the ten-paraphrase robustness setting of
VisualSwap \citep{shi2026visualswap}. Each variant conveys the same
instruction to re-examine the current visual input but uses a distinct
lexical realization. The strings below are independently worded,
protocol-matched realizations of the same ten prompt intents rather
than verbatim reproductions. For every eligible model--sample pair, only
\(P_{\mathrm{reflect}}\) in Eq.~(\ref{eq:history-state}) is replaced;
the current image, question, prior CoT, assistant-turn structure,
decoding configuration, and evaluator are held fixed. We evaluate the
same paired cohort under all ten variants and report the mean and
standard deviation across prompts for VUR, PAR, and the
answer-preference margin \(m\). To separately test the semantics of the
reflection prefix, we replace \(P_{\mathrm{reflect}}\) with the neutral
assistant continuation ``Let me continue the reasoning.'' while
holding all other inputs and settings fixed.

\begin{table}[t]
\centering
\scriptsize
\setlength{\tabcolsep}{3pt}
\renewcommand{\arraystretch}{1.04}
\begin{tabular}{@{}cp{0.84\columnwidth}@{}}
\toprule
\textbf{ID} & \textbf{Reflection prompt} \\
\midrule
1 & Let me inspect the image more closely before accepting this conclusion. \\
2 & I should revisit the figure and verify each step of my reasoning. \\
3 & Let me examine the smallest visual details once more before answering. \\
4 & I will check the visual input again to confirm that observation. \\
5 & Let me reassess the complete figure and confirm that my reading is consistent. \\
6 & I should focus again on the relevant local region and verify this point. \\
7 & Let me carefully re-inspect the figure before relying on my first impression. \\
8 & I will take another look at the image in case I interpreted it incorrectly. \\
9 & Let me verify the visual evidence again to rule out a perception error. \\
10 & I should inspect the image once more and ensure the claimed detail is actually present. \\
\bottomrule
\end{tabular}
\caption{\textbf{Reflection-prompt variants for the robustness
evaluation.} Following VisualSwap, the variants preserve the intent of
visual re-examination while varying its lexical expression.}
\label{tab:reflection-prompt-variants}
\end{table}

\subsection{Naturally Occurring Errors}
\label{app:natural-errors}

We additionally run standard inference on the original benchmark
images without an image swap and retain model--sample pairs whose
initial generated answer is incorrect. Eligibility depends only on
this initial error and does not require the prompt-only condition to
recover the correct answer. The ordinary-reflection condition retains
the model-generated erroneous CoT, whereas the matched prompt-only
condition removes that CoT while preserving the image, question,
reflection instruction, decoding configuration, and evaluator. We
report the percentage of responses corrected to the benchmark target
under each condition.

\section{FSAF Implementation and Additional Analysis}
\label{app:fsaf-details}

\subsection{Exact Re-Query and Execution Procedure}
\label{app:fsaf-execution}

FSAF uses the same fixed user turn for every model and sample:
\begin{quote}
\small
\emph{Answer the same question using only the current image. Be concise
and end with the final-answer format requested by the question.}
\end{quote}
The intervention then proceeds as follows. First, the model generates
the prior assistant CoT \(R^{-}\), while the serving runtime records its
token interval \([s^{-},e^{-})\). Second, the fixed re-query \(U_f\) is
appended as a new user turn. Third, the runtime constructs
\(M^{\mathrm{FSAF}}\) in Eq.~(\ref{eq:fsaf-mask}) directly from the
recorded interval, without parsing the answer or locating individual
evidence spans. Fourth, the mask is enabled in every language layer for
both the prefill of \(U_f\) and the generation of \(R^{F}\). Finally,
the model produces one response using the same decoding settings and
maximum generation length as the baseline. Thus, FSAF adds neither a
second model call nor a candidate-answer extraction stage.

\subsection{Implementation and Deployment Scope}
\label{app:fsaf-implementation}

Our implementation applies the additive mask through
Transformers eager-attention hooks for Qwen2.5-VL and Qwen3-VL. The
runtime identifies fresh-suffix queries by their positions after
\(e^{-}\) and masks all keys in \([s^{-},e^{-})\). The intervention
covered all 28 language layers of Qwen2.5-VL-7B, all 36 layers of the
Qwen3-VL-8B variants, and all 64 layers of the Qwen3-VL-32B variants.
The current implementation uses batch size \(1\); a vLLM deployment
requires the same additive mask to be integrated into its attention
backend or a custom operator. A fail-closed policy enables both the
re-query and firewall only for the five verified model configurations.
For an unknown model, both components are disabled and the caller's
ordinary reflection baseline is used. This avoids silently degrading
to the re-query-only condition.

\subsection{Design Ablations}
\label{app:fsaf-ablation}

We study why FSAF needs both a fresh user--assistant boundary and a
firewall spanning prefill and generation. Tab.~\ref{tab:fsaf-ablation}
reports the Qwen2.5-VL-7B design study. Re-answer querying (RAQ) masks
only generated-token access to \(R^{-}\); the re-query cue can therefore
read \(R^{-}\) during prefill and carry it forward in its cached states.
Clean recheck framing (CRF) also cleans the cue, but continues within
the same assistant turn. The re-query-only condition uses the exact
FSAF wording without changing attention accessibility.

\begin{center}
\centering
\scriptsize
\setlength{\tabcolsep}{2.8pt}
\renewcommand{\arraystretch}{1.10}
\begin{tabular}{@{}llcc@{}}
\toprule
\textbf{Condition}
& \textbf{Evaluator}
& \multicolumn{2}{c}{\textbf{vs. Ordinary Reflection (pp)}}\\
\cmidrule(lr){3-4}
&
& \(\Delta\)\textbf{VUR \(\uparrow\)}
& \(\Delta\)\textbf{PAR \(\downarrow\)}\\
\midrule
RAQ: generation-only mask
& 235B judge
& \(-2.62\)
& \(-35.21\)\\
CRF: same-turn cue cleaning
& Rule
& \(+0.75\)
& \(-16.85\)\\
Re-query only
& Rule
& \shortstack{\(+3.00\)\\{\tiny \([-2.25,+8.24]\)}}
& \shortstack{\(+19.48\)\\{\tiny \([+11.99,+26.97]\)}}\\
Full FSAF
& Rule
& \shortstack{\(\mathbf{+23.97}\)\\{\tiny \([+17.60,+30.34]\)}}
& \shortstack{\(\mathbf{-14.61}\)\\{\tiny \([-20.98,-7.87]\)}}\\
Full FSAF
& 235B judge
& \shortstack{\(\mathbf{+18.73}\)\\{\tiny \([+12.36,+25.09]\)}}
& \shortstack{\(\mathbf{-31.46}\)\\{\tiny \([-37.83,-25.09]\)}}\\
\bottomrule
\end{tabular}
\captionof{table}{\textbf{FSAF design ablations on Qwen2.5-VL-7B.}
Values are paired percentage-point differences relative to ordinary
reflection; brackets give \(95\%\) confidence intervals when reported.
Rule-based rows are auxiliary design screens, whereas the final paper
claim uses the 235B semantic judge.}
\label{tab:fsaf-ablation}
\end{center}

RAQ substantially reduces PAR but also decreases VUR, showing that
removing a direct generation-time edge does not by itself preserve
correct updating. The rule-based re-query-only screen yields no reliable
VUR gain and increases PAR, while full FSAF improves VUR and reduces
PAR under both the rule screen and the 235B judge. Because CRF and
re-query-only were not evaluated with the final semantic judge in this
package, we use them only to motivate the coupled design, rather than
as standalone semantic-effect claims.

\subsection{State-Conditioned Transitions}
\label{app:fsaf-transitions}

To identify which baseline outcomes contribute to the pooled effect in
Tab.~\ref{tab:fsaf-main}, we stratify each paired instance by the 235B
judge label assigned to its ordinary-reflection output. \emph{Recovered}
denotes a baseline output matching only \(y\), \emph{Active stale}
denotes one matching only \(y^{-}\), and \emph{Both} and \emph{Other}
denote outputs matching both or neither target, respectively. We then
measure the paired FSAF outcome within each fixed baseline stratum.

\begin{table}[t]
\centering
\scriptsize
\setlength{\tabcolsep}{1.4pt}
\renewcommand{\arraystretch}{1.12}
\begin{tabular}{@{}lrrr@{}}
\toprule
\shortstack{\textbf{Baseline}\\\textbf{outcome}}
& \textbf{Pairs}
& \textbf{VUR: Base \(\rightarrow\) FSAF}
& \textbf{PAR: Base \(\rightarrow\) FSAF}\\
\midrule
Recovered
& 625
& \(100.00 \rightarrow 82.40\;(-17.60)\)
& \(0.00 \rightarrow 1.12\;(+1.12)\)\\
Active stale
& 696
& \(0.00 \rightarrow 38.51\;(+38.51)\)
& \(100.00 \rightarrow 6.47\;(-93.53)\)\\
Other
& 469
& \(0.00 \rightarrow 37.74\;(+37.74)\)
& \(0.00 \rightarrow 2.77\;(+2.77)\)\\
Both
& 10
& \(100.00 \rightarrow 50.00\;(-50.00)\)
& \(100.00 \rightarrow 10.00\;(-90.00)\)\\
\bottomrule
\end{tabular}
\caption{\textbf{FSAF transitions conditioned on the ordinary-reflection
outcome.} Each metric cell reports baseline \(\rightarrow\) FSAF,
followed by the paired change in parentheses. Rates and changes are in
percent and percentage points, respectively.}
\label{tab:fsaf-state-transitions}
\end{table}

The pooled improvement is concentrated in the \emph{Active stale} and
\emph{Other} strata. For active-stale outputs, FSAF reduces PAR from
\(100.00\%\) to \(6.47\%\) and raises VUR to \(38.51\%\); among Other
outputs, \(37.74\%\) become current-answer matches. FSAF preserves the
current answer for \(82.40\%\) of recovered outputs, showing that the
intervention does not perfectly preserve already successful
reflections. At the pooled level, the Other share rises from
\(26.06\%\) to \(43.17\%\), so lower PAR does not imply that every
released stale answer becomes the current answer. Because these strata
are defined by the observed baseline
output, the values describe state-conditioned transitions rather than
causal differences between latent states. We do not interpret the
\emph{Both} stratum further because it contains only 10 pairs.

\subsection{Evaluation and Intervention Audit}
\label{app:fsaf-audit}

The final semantic evaluation contains \(3{,}600/3{,}600\) expected
baseline/FSAF judge tasks and \(1{,}800/1{,}800\) paired model--sample
instances across the five models, with no missing or duplicate task
IDs. The answer-preference evaluation contains \(5{,}400/5{,}400\)
finite scoring records across baseline, re-query-only, and FSAF, forming
\(1{,}800/1{,}800\) complete three-condition model--sample triplets with
no duplicates. All FSAF generations have at least \(0.90\) recorded
coverage of the intended \(R^{-}\) span, and every sample records
modified attention edges during both prefill and generation. Baseline
and FSAF use the same maximum generation length of 256 tokens. The rule
evaluator and 235B judge disagree on at least one VUR/PAR label for
\(951/3{,}600\) outputs (\(26.42\%\)); consequently, the VUR/PAR columns
in Tab.~\ref{tab:fsaf-main} use the 235B semantic judge, while its
answer-preference columns use the evaluated models' mean token
log-probabilities. Rule-based results serve only as low-cost design
screens.

\section{VS-Bench Construction and Use}
\label{app:vsbench}

\noindent\textbf{Scope and Composition.}
VS-Bench \citep{shi2026visualswap} is designed to test visual
re-examination rather than static visual question answering. It contains
\(800\) image pairs, with \(200\) pairs drawn from each of
MathVista-MINI, the Vision-Dominant subset of MathVerse-MINI,
MathVision, and MMMU-Pro
\citep{lu2024mathvista,zhang2024mathverse,wang2024mathvision,
yue2025mmmupro}. These sources span charts, function plots, geometry,
competition mathematics, and expert-level multimodal questions, while
retaining questions whose answers depend on visual evidence.

\noindent\textbf{Pair Construction.}
Each instance contains a shared question \(Q\), a source benchmark image
\(I_b\), and a human-checked alternative image \(I_a\). The construction
follows three constraints: (i) \(Q\) remains valid for both images;
(ii) the images preserve high-level layout, style, composition, and
context; and (iii) an answer-critical visual detail changes, yielding
distinct ground-truth answers \(A_a\neq A_b\). VisualSwap uses a
human-in-the-loop image-editing process: annotators identify the
answer-critical element, specify a targeted modification, and iteratively
refine the edited image. The paired images are also kept at identical
resolution so that the swap does not alter the visual-token count.

\begin{table}[h!]
\centering
\small
\setlength{\tabcolsep}{5pt}
\begin{tabular}{lcccc}
\toprule
\textbf{Source} & \textbf{Pairs}
& \textbf{CLIP \(\uparrow\)}
& \textbf{SSIM \(\uparrow\)}
& \textbf{LPIPS \(\downarrow\)} \\
\midrule
MathVista-MINI  & 200 & 0.95 & 0.88 & 0.12 \\
MathVerse-MINI  & 200 & 0.97 & 0.94 & 0.06 \\
MathVision      & 200 & 0.95 & 0.86 & 0.17 \\
MMMU-Pro        & 200 & 0.94 & 0.75 & 0.19 \\
\midrule
Overall         & 800 & 0.95 & 0.86 & 0.14 \\
\bottomrule
\end{tabular}
\caption{\textbf{VS-Bench composition and image-pair similarity.}
Higher CLIP similarity and SSIM, together with lower LPIPS, indicate
that paired images remain visually similar despite their
answer-determining differences. Values are reported by VisualSwap
\citep{shi2026visualswap}.}
\label{tab:vsbench-composition}
\end{table}

\noindent\textbf{Use in Our Framework.}
To match the notation in our controlled framework, we relabel the source
benchmark image as the current image \(I=I_b\) and the edited alternative
as the counterfactual image \(I^{-}=I_a\), with targets \(y=A_b\) and
\(y^{-}=A_a\). This direction lets the evaluated model first produce
\(R^{-}\) on the coherent counterfactual task and then reflect under the
current benchmark image. The full \(800\)-pair benchmark defines the
available task pool; each model's analyzed cohort is subsequently
restricted to pairs that it answers correctly under direct inference on
both \(I^{-}\) and \(I\). Thus, the controlled comparison targets
prior-CoT influence rather than failures caused by an independently
unsolved image.

\section{Cohort Construction, Sample Counts, and Coverage}
\label{app:cohort-coverage}

\subsection{Eligibility and Sample Flow}
\label{app:eligibility-flow}

The available pool contains \(800\) paired tasks per model. A
model--sample pair enters the corrected diagnosis cohort only when the
model's direct responses to both \(I^{-}\) and \(I\) semantically match
their respective targets under the 235B judge. This model-specific
filter prevents an unsolved source or current task from being
misclassified as resistance to, or persistence of, a prior state.
Eligibility is therefore defined at the model--sample level rather than
once for the benchmark.

The unified 16-model diagnosis contains \(4{,}159\) eligible
model--sample pairs. Subsequent analyses retain this identity key and
apply only the filter required by their estimand. For example, the
matched support-withdrawal test selects the \(2{,}456\) instances that
first recover the current answer under ordinary reflection.

The FSAF evaluation uses the original rule-extracted eligibility
snapshot frozen for the intervention runs. It requires an answer-changing
pair for which direct inference is correct on both images. In the model
order of Tab.~\ref{tab:fsaf-main}, its counts are
\(267/420/486/322/305\), totaling \(1{,}800\), rather than a
\(360\)-per-model subsample. It is not a post-hoc filter of the
\(1{,}943\) five-Qwen pairs in Tab.~\ref{tab:eligible-counts}; those
counts \((258/414/469/422/380)\) come from the corrected 235B
semantic-judge eligibility snapshot used by the 16-model diagnosis.
Because the evaluator and membership differ, all FSAF effects remain
within the fixed \(1{,}800\)-pair cohort. The generated baseline and
FSAF outputs are still evaluated by the 235B semantic judge, as detailed
in Appendix~\ref{app:fsaf-audit}.

\begin{table}[t]
\centering
\scriptsize
\setlength{\tabcolsep}{4pt}
\renewcommand{\arraystretch}{1.10}
\begin{tabular}{@{}p{0.48\columnwidth}rp{0.28\columnwidth}@{}}
\toprule
\textbf{Stage or analysis} & \textbf{Pairs} & \textbf{Additional filter} \\
\midrule
Available task pool
& \(16\times800\) & None \\
Corrected 16-model diagnosis
& \(4{,}159\) & Directly correct on both images \\
Matched support withdrawal
& \(2{,}456\) & Recovered under ordinary reflection \\
FSAF semantic evaluation
& \(1{,}800\) & Five verified Qwen configurations \\
FSAF judge tasks
& \(3{,}600\) & Baseline and FSAF output per pair \\
FSAF preference records
& \(5{,}400\) & Baseline, re-query-only, and FSAF per pair \\
\bottomrule
\end{tabular}
\caption{\textbf{Sample flow for the principal analysis cohorts.}
Counts are model--sample pairs, not unique benchmark questions.}
\label{tab:sample-flow}
\end{table}

\subsection{Eligible Counts by Model}
\label{app:eligible-counts}

Tab.~\ref{tab:eligible-counts} reports the corrected semantic-judge
eligibility count for every checkpoint. These counts explain why
unweighted cross-model raw rates can be misleading and motivate the
paired estimand in Appendix~\ref{app:paired-effects}. They are not
performance scores: a model can have a smaller eligible cohort for many
reasons, including failure on either direct-image task.

\begin{table}[t]
\centering
\scriptsize
\setlength{\tabcolsep}{4pt}
\renewcommand{\arraystretch}{1.05}
\begin{tabular}{@{}lr@{}}
\toprule
\textbf{Model} & \textbf{Eligible pairs} \\
\midrule
Qwen2.5-VL-7B-Instruct & 258 \\
Qwen3-VL-8B-Instruct & 414 \\
Qwen3-VL-8B-Thinking & 469 \\
Qwen3-VL-32B-Instruct & 422 \\
Qwen3-VL-32B-Thinking & 380 \\
InternVL2-8B & 141 \\
Gemma-4-12B & 60 \\
Gemma-4-12B-it & 229 \\
Gemma-4-31B & 136 \\
Gemma-4-31B-it & 275 \\
Qwen3.5-4B & 233 \\
Qwen3.5-9B & 242 \\
Qwen3.5-27B & 267 \\
Qwen3.6-27B & 269 \\
Kimi-VL-A3B-Instruct & 181 \\
Kimi-VL-A3B-Thinking-2506 & 183 \\
\midrule
\textbf{Total} & \textbf{4,159} \\
\bottomrule
\end{tabular}
\caption{\textbf{Corrected eligible model--sample pairs in the
16-model diagnosis.} Eligibility requires semantic correctness under
direct inference on both paired images.}
\label{tab:eligible-counts}
\end{table}

\subsection{Completeness and Coverage Checks}
\label{app:coverage-checks}

Before aggregation, each analysis verifies unique task IDs and the
presence of every condition required for a matched comparison. The
FSAF semantic package contains \(3{,}600/3{,}600\) expected judge
tasks and \(1{,}800/1{,}800\) complete baseline/FSAF pairs. Its
answer-preference package contains \(5{,}400/5{,}400\) finite records
and \(1{,}800/1{,}800\) complete three-condition triplets, with no
duplicate task IDs. All FSAF rows record at least \(0.90\) coverage of
the intended stale span and nonzero modified attention edges in both
prefill and generation. Experiment-specific subsets are reported with
their own denominator; missing conditions are never imputed.

\section{Representative Case Studies}
\label{app:case-studies}

Figs.~\ref{fig:case-s1}--\ref{fig:case-p11} visualize one frozen,
representative model--sample pair for each principal diagnostic or
continuation experiment. Following the presentation style of
VisualSwap \citep{shi2026visualswap}, each figure reads from the prior
observation \((I^{-},Q,R^{-})\), through the current image \(I\), to the
condition-specific intervention and observed response. Red denotes the
prior state and green denotes the current state. For candidate scoring,
the displayed margin is
\(\log p(y\mid C)-\log p(y^{-}\mid C)\), so positive values favor the
current answer; generation labels are assigned separately by the
semantic judge. Long questions or prior responses are excerpted with
ellipses only for display; the experiment uses the complete strings.
These cases illustrate the protocol and observed behaviors rather than
serving as frequency estimates.

\subsection{Diagnosing Prior-State Control}

\begin{figure*}[t]
\centering
\includegraphics[width=\textwidth]{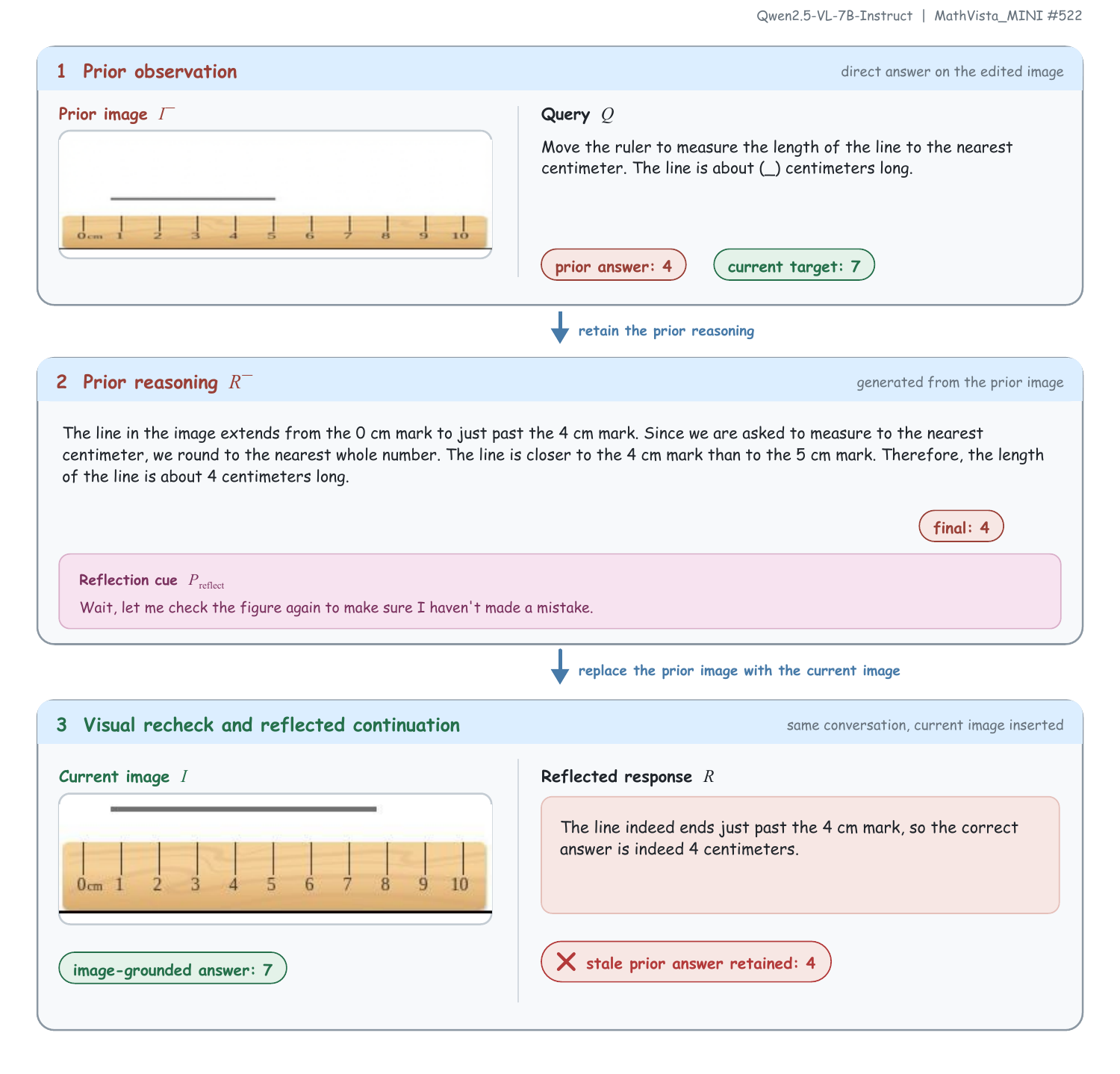}
\caption{\textbf{A prior CoT can override changed visual evidence.}
Top: Qwen2.5-VL-7B-Instruct answers \(y^{-}=4\) from the prior image
\(I^{-}\) and records the supporting response \(R^{-}\).
Middle: the image is replaced by the current image \(I\), whose target
is \(y=7\), while the prior response remains in the conversation.
Bottom: after the reflection cue, the model repeats the prior answer
instead of recomputing from the changed ruler. This representative
failure illustrates the competition between fresh visual recomputation
and stale textual reuse described in the main paper.}
\label{fig:case-s1}
\end{figure*}

\begin{figure*}[t]
\centering
\includegraphics[width=\textwidth]{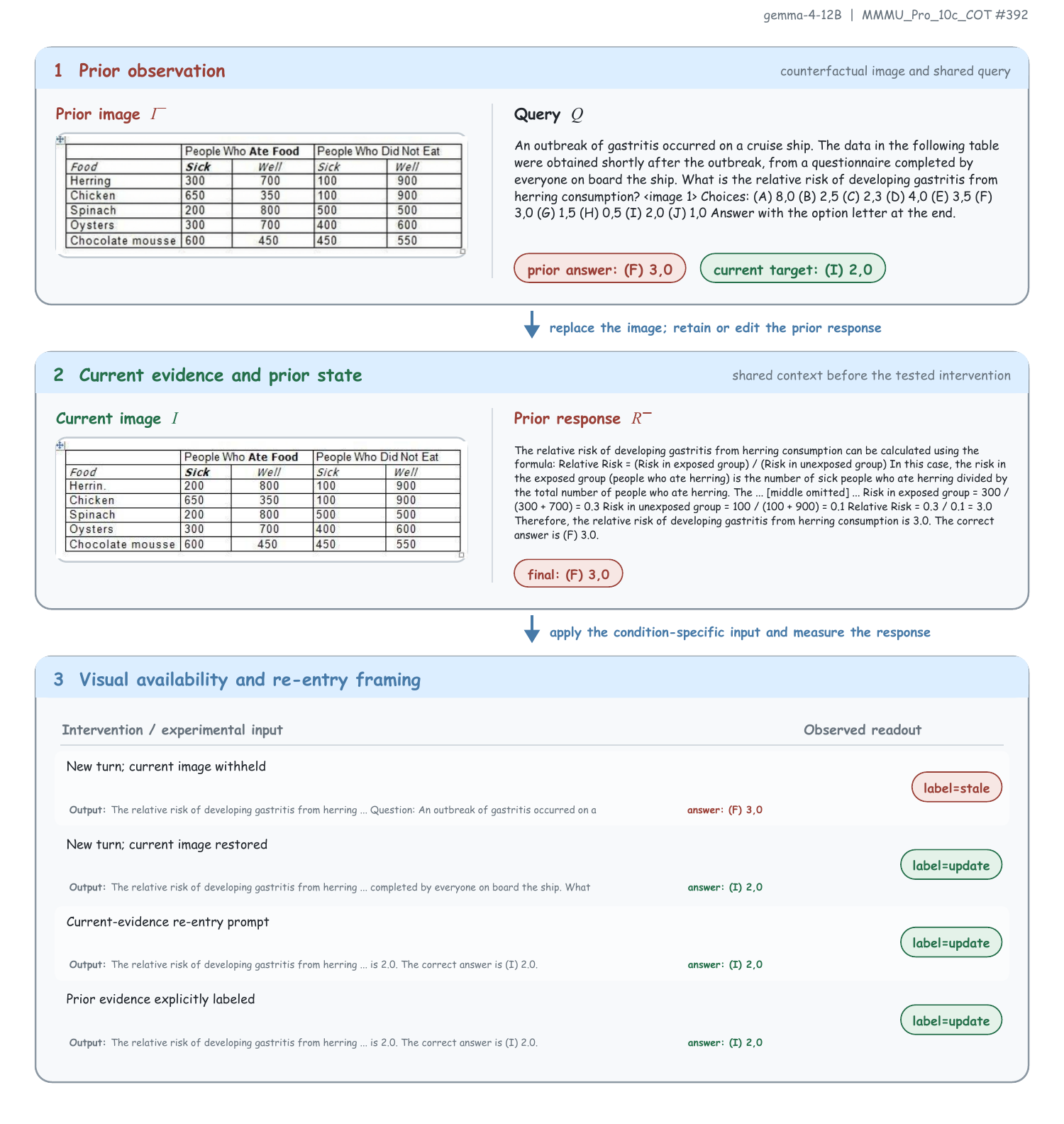}
\caption{\textbf{Current visual evidence can reopen recomputation.}
Top and middle: Gemma-4-12B first derives the prior relative risk
\(y^{-}=3.0\), after which the current table changes the target to
\(y=2.0\).
Bottom: a fresh turn without the current image remains stale, whereas
restoring \(I\), explicitly requesting current-evidence re-entry, or
labeling the carried evidence as prior all yield the current answer.
The contrast illustrates the main-paper finding that \(R^{-}\) can
sustain a competing answer path when \(I\) is absent, while restoring
and foregrounding current visual evidence can shift the model back
toward recomputation.}
\label{fig:case-p01}
\end{figure*}

\begin{figure*}[t]
\centering
\includegraphics[width=\textwidth]{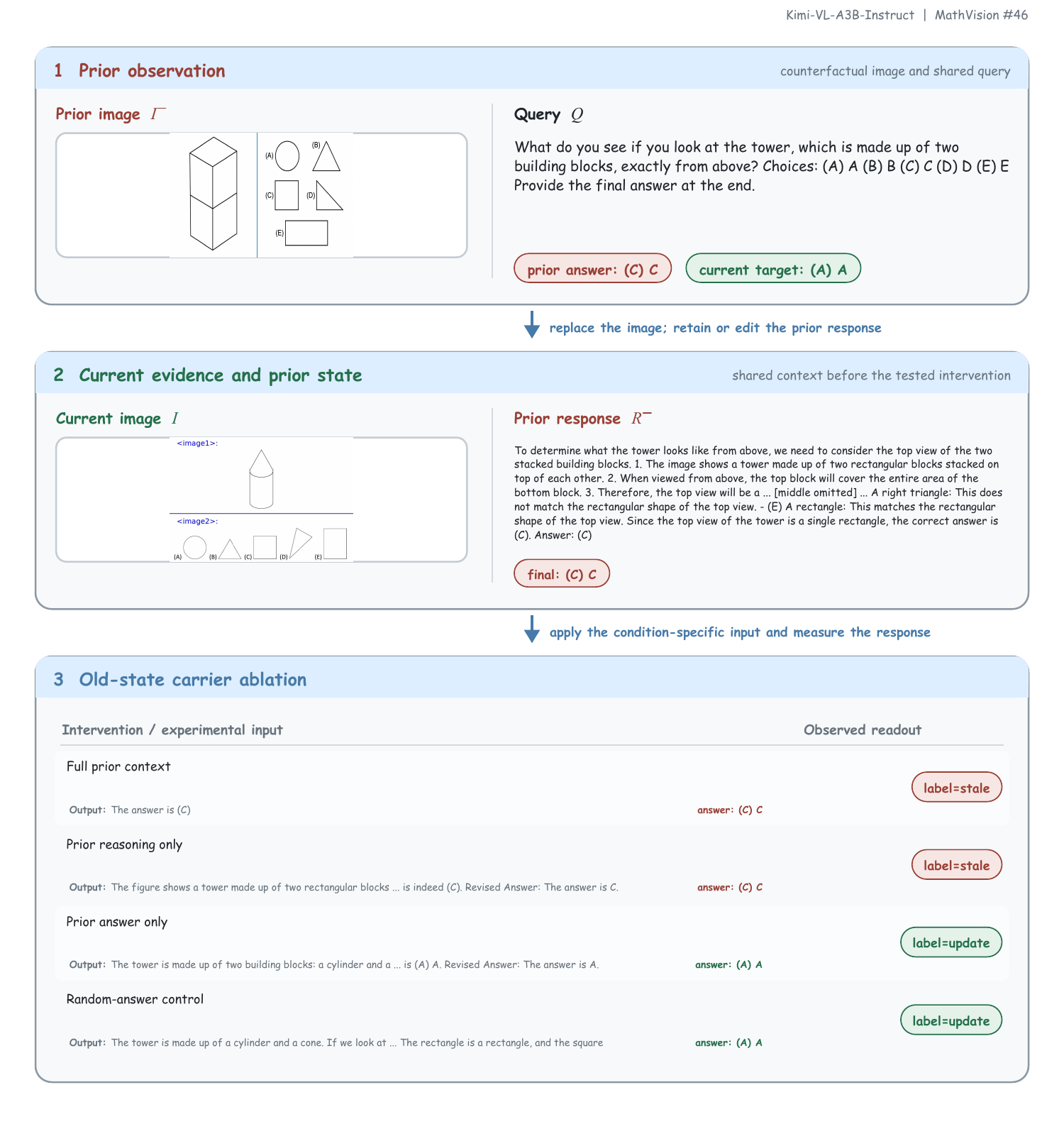}
\caption{\textbf{Reasoning, not the answer token alone, carries
prior-state control.}
The current geometry image \(I\) is fixed across all four conditions.
Kimi-VL-A3B-Instruct remains stale when given either the full prior
response or its reasoning body alone, but updates when only the prior
answer or a random answer control is carried forward. In this
representative contrast, task-relevant reasoning in \(R^{-}\), rather
than the answer token by itself, provides the reusable route to the
prior answer.}
\label{fig:case-s2}
\end{figure*}

\begin{figure*}[t]
\centering
\includegraphics[width=\textwidth]{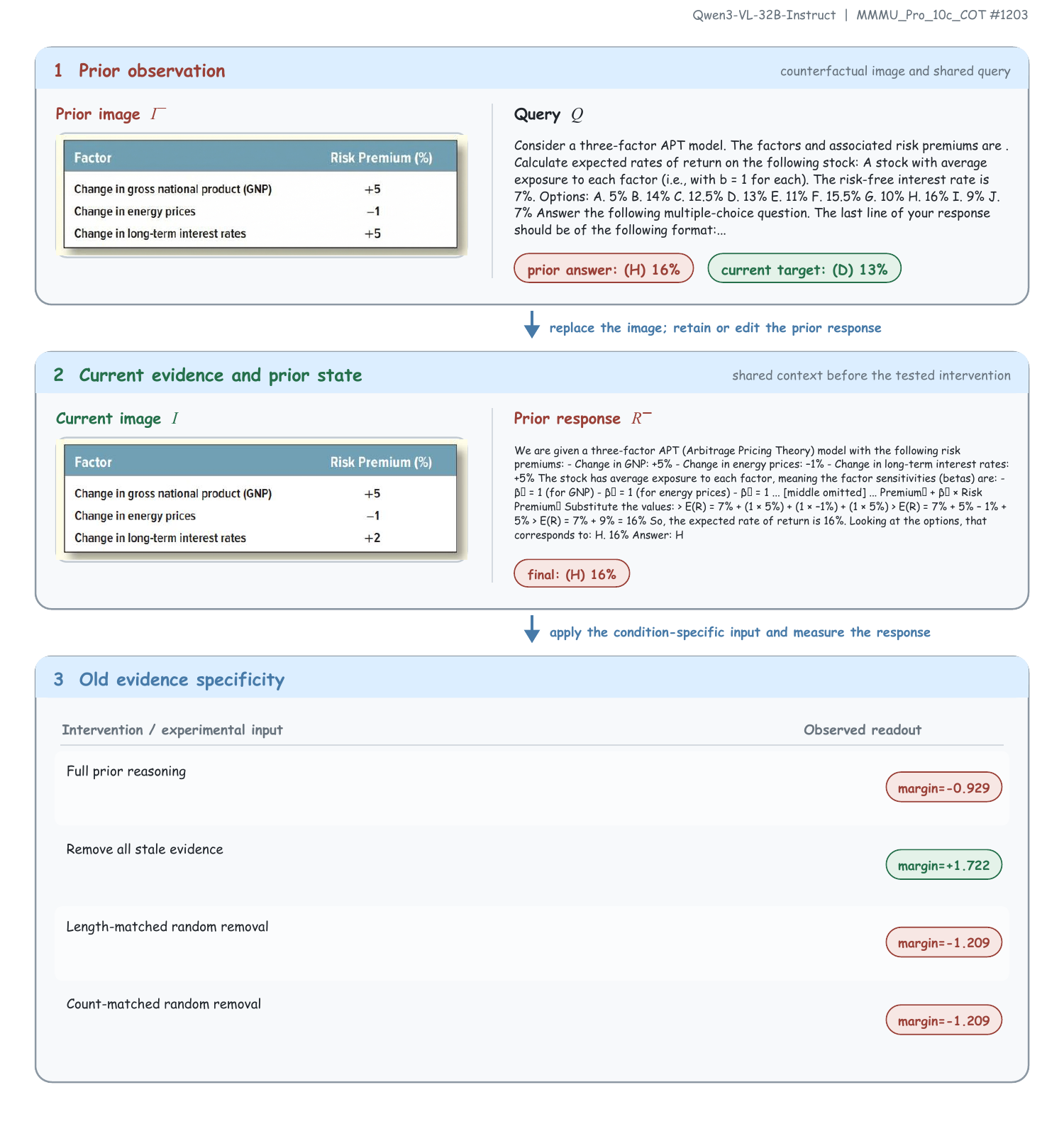}
\caption{\textbf{Evidence-bearing content carries textual-shortcut
control.}
Qwen3-VL-32B-Instruct initially favors the prior answer when the full
prior response is retained. Removing all answer-supporting evidence
switches the candidate margin to the current answer, whereas removing a
length-matched or count-matched random span does not. This matched case
illustrates the content-specific effect reported in the main paper:
weakening the stale evidence route, rather than merely shortening the
prefix, reduces prior-answer control.}
\label{fig:case-p03}
\end{figure*}

\begin{figure*}[t]
\centering
\includegraphics[width=\textwidth]{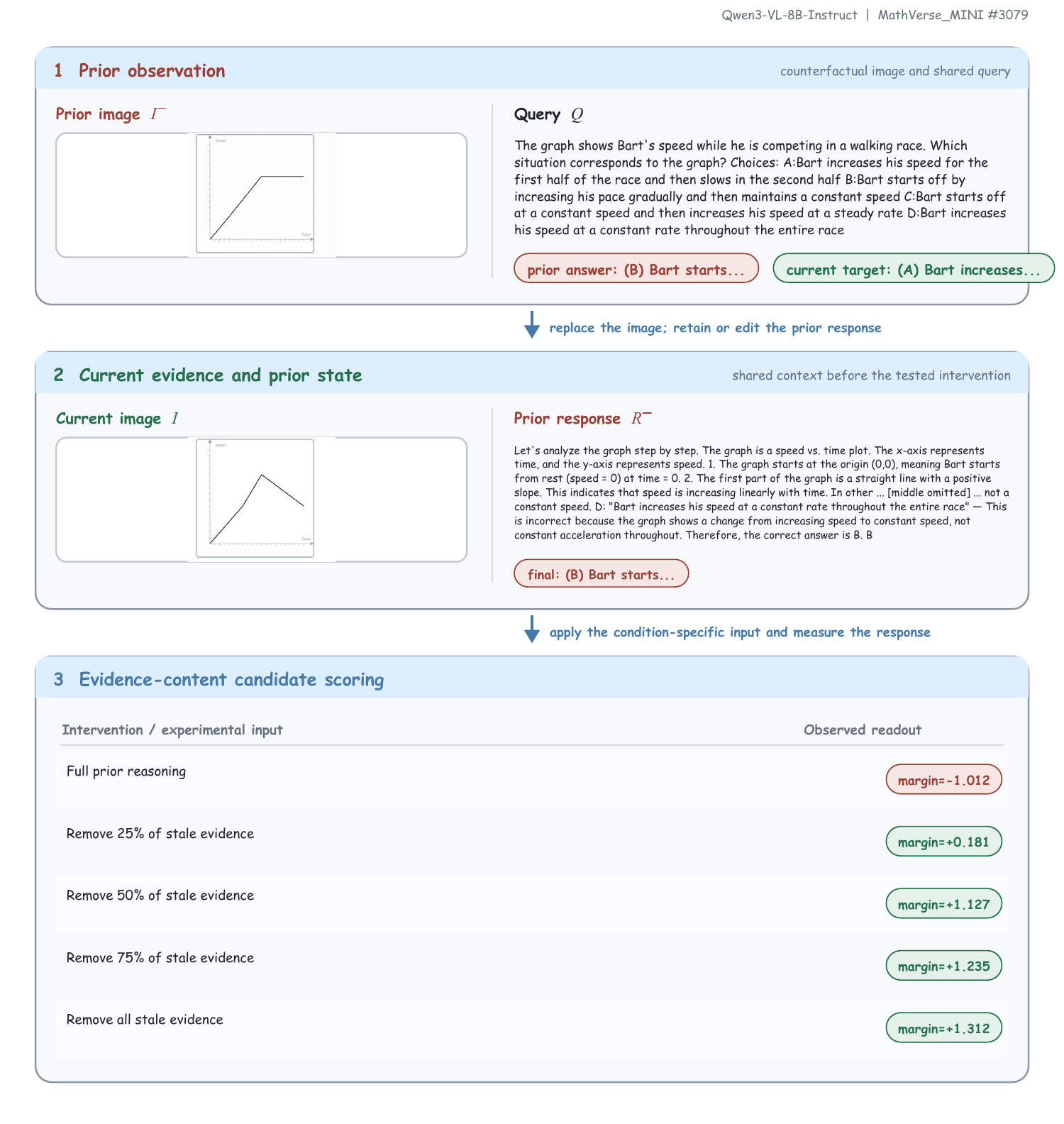}
\caption{\textbf{Evidence removal progressively weakens prior-answer
control.}
Qwen3-VL-8B-Instruct begins with a prior-favored margin under the full
response \(R^{-}\). As \(25\%\), \(50\%\), \(75\%\), and then all of the
identified stale-evidence spans are removed, the margin progressively
moves toward and then favors the current answer. The case visualizes
the graded weakening of textual-shortcut control as the
answer-supporting route in \(R^{-}\) is withdrawn.}
\label{fig:case-s6s}
\end{figure*}

\begin{figure*}[t]
\centering
\includegraphics[width=\textwidth]{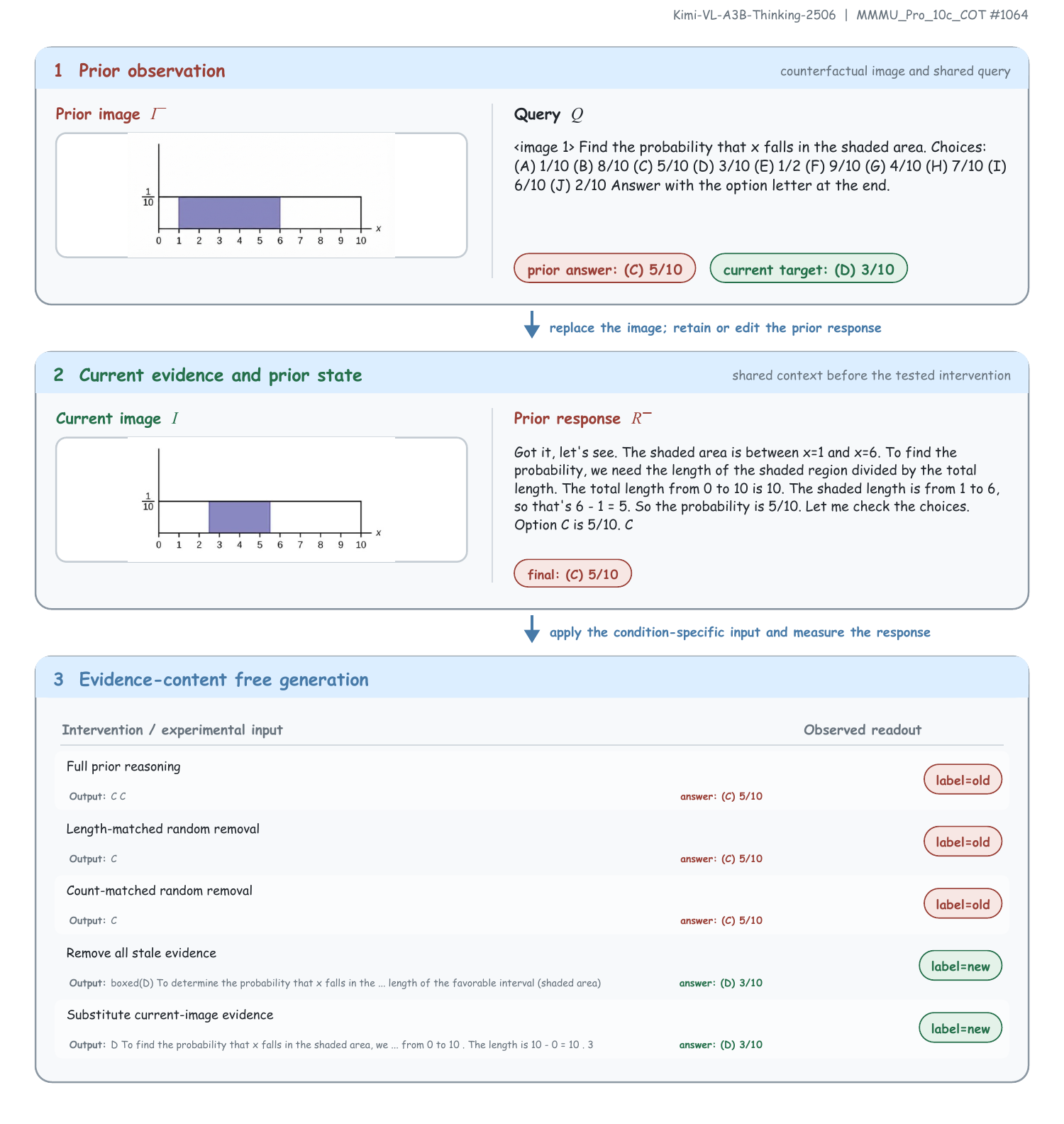}
\caption{\textbf{Evidence-bearing content also controls free
generation.}
With the current chart fixed, Kimi-VL-A3B-Thinking-2506 repeats the
prior answer under the full \(R^{-}\) and after a length-matched random
removal. Removing all stale evidence or replacing it with current-image
evidence yields the current answer. This generated-output contrast
supports the same main-paper conclusion as candidate scoring:
evidence-bearing content is the robust carrier of the textual
shortcut.}
\label{fig:case-s6g}
\end{figure*}

\begin{figure*}[t]
\centering
\includegraphics[width=\textwidth]{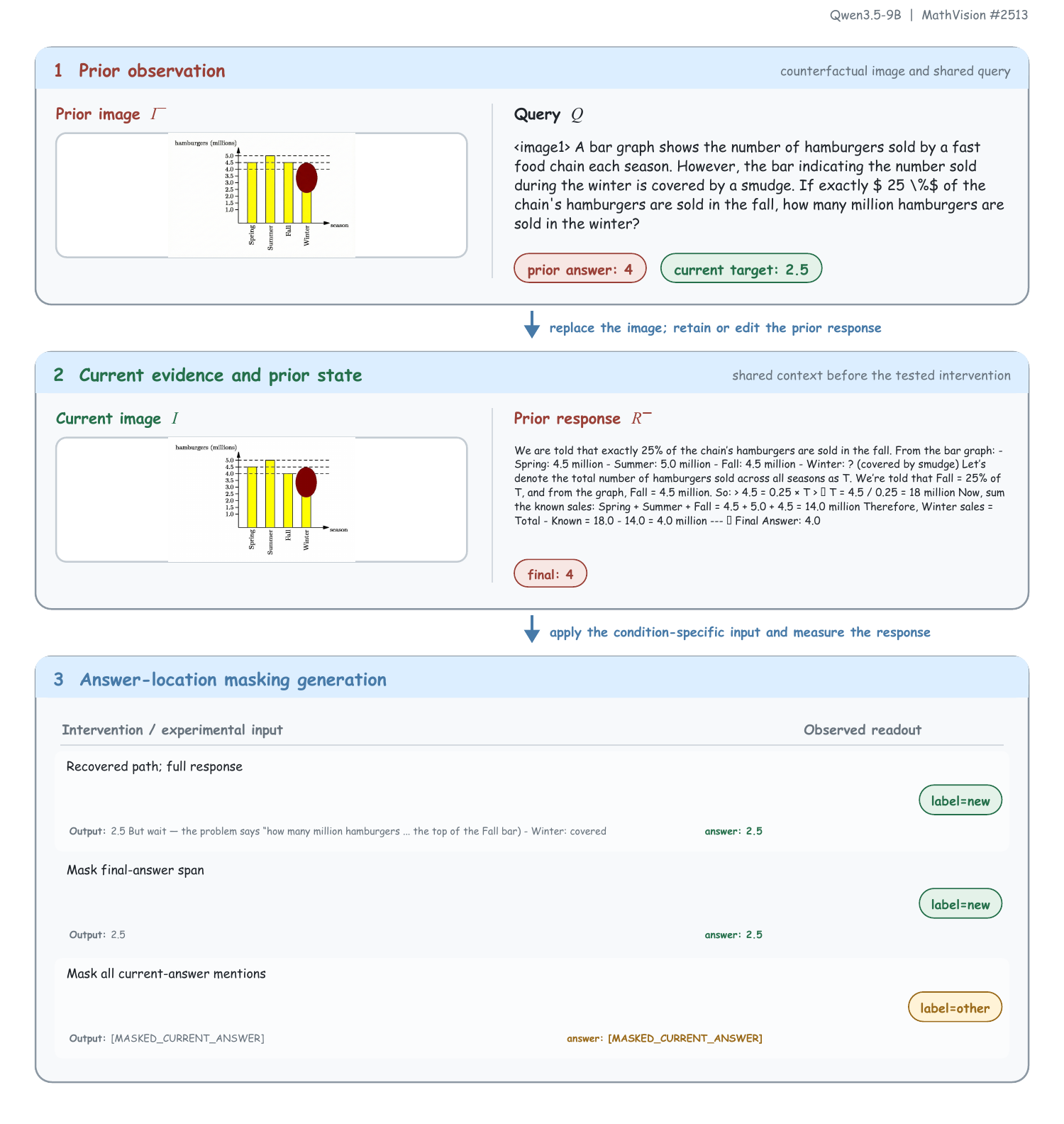}
\caption{\textbf{A recovered response can carry current-answer
information beyond its final span.}
Qwen3.5-9B remains current-correct after the final answer span is
masked, but no longer produces the current answer after all mentions of
that answer within the recovered response are masked. For this case,
recovery is therefore not stored only in the terminal answer string;
answer-relevant information distributed through the current reasoning
also matters.}
\label{fig:case-s6m}
\end{figure*}

\begin{figure*}[t]
\centering
\includegraphics[width=\textwidth]{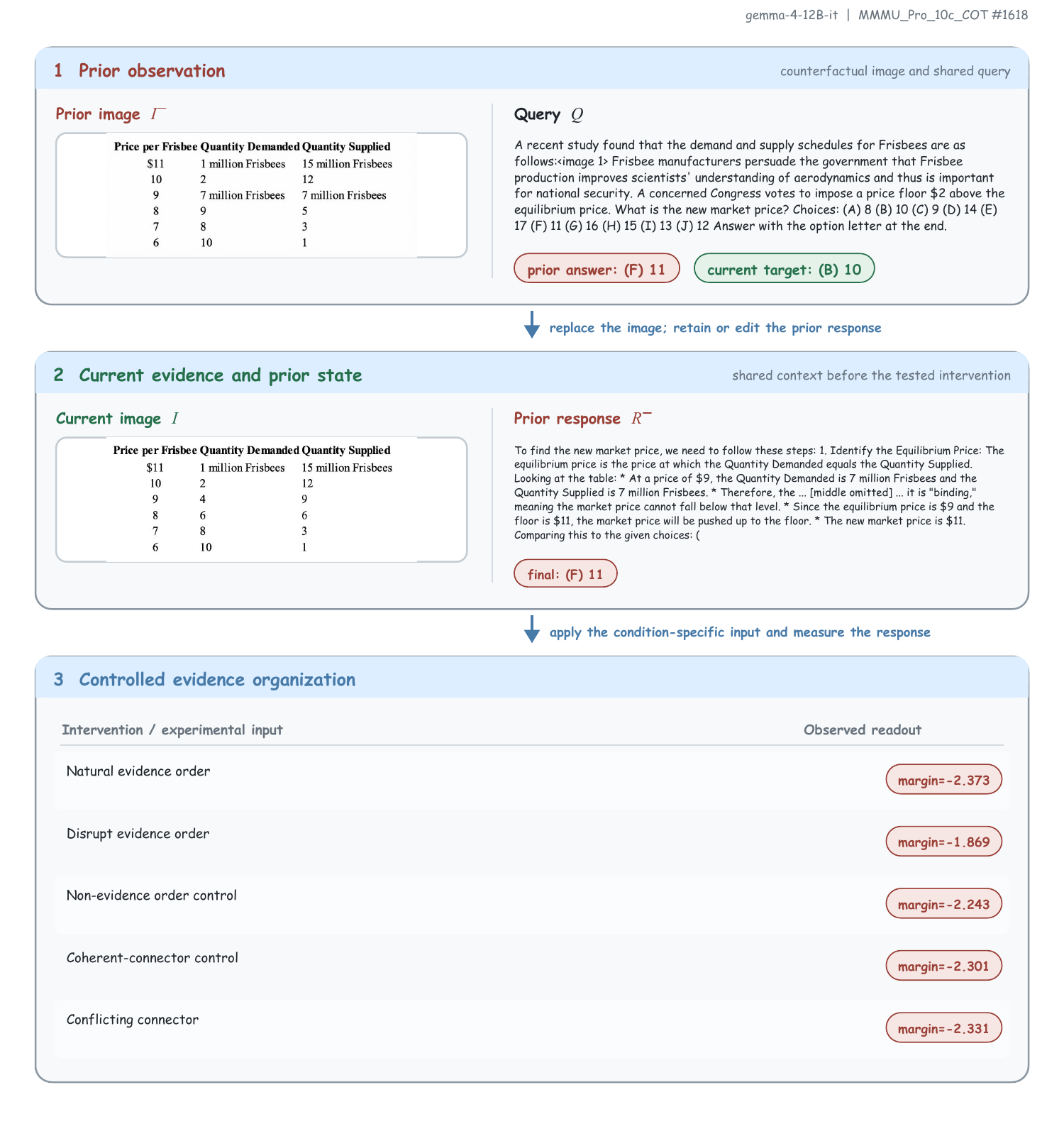}
\caption{\textbf{Disrupting evidence-chain organization weakens
textual-shortcut control.}
Gemma-4-12B-it favors the prior answer under the natural response.
Disrupting the order of evidence-bearing sentences weakens this
preference more than either disrupting non-evidence sentences or
editing connective wording, although the case remains prior-favored.
This controlled example mirrors the aggregate result that logical
organization strengthens reuse of evidence-bearing content, while
remaining a moderator rather than the shortcut's sole cause.}
\label{fig:case-p15}
\end{figure*}

The first group traces prior-state control from the basic visual-recheck
failure to its carrier, evidence content, strength, and organization.
Within each controlled contrast, the current image and query are fixed
while the retained or edited prior response changes. Reading
Figs.~\ref{fig:case-s1}--\ref{fig:case-p15} in sequence therefore shows
which parts of \(R^{-}\) sustain the prior-answer path and which
interventions reopen current-image recomputation.

\subsection{Testing Shortcut Retirement and Future Transfer}

\begin{figure*}[t]
\centering
\includegraphics[width=\textwidth]{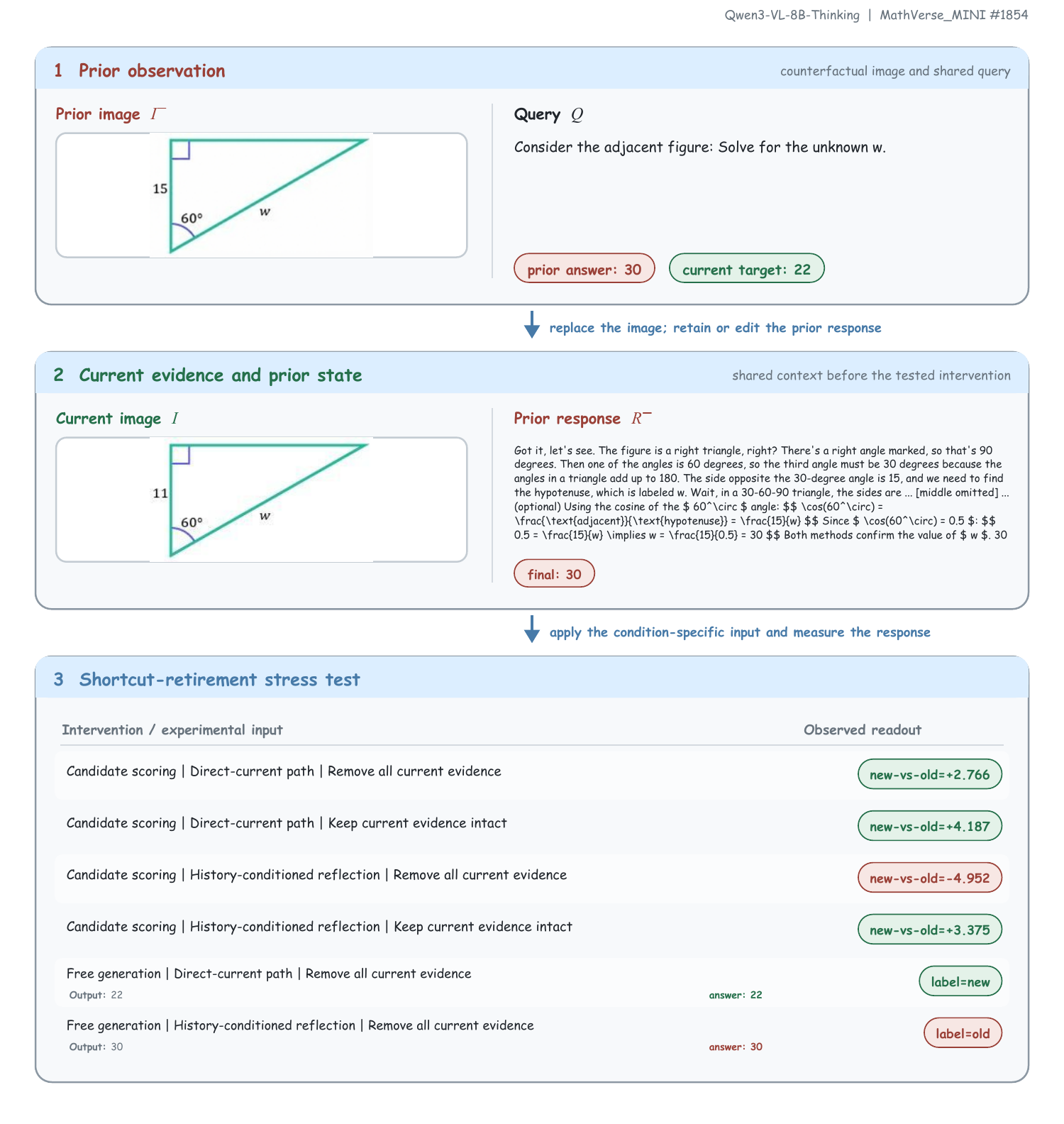}
\caption{\textbf{Matched support withdrawal reveals prior-history
dependence after answer recovery.}
Qwen3-VL-8B-Thinking has recovered the current answer before this test.
When current supporting evidence is removed, the direct-current path
remains current-favored, but the history-conditioned path relapses to a
prior-favored candidate margin; both paths favor the current answer
when that evidence remains intact. Candidate scoring and free
generation are shown as separate readouts. The case illustrates why a
correct reflected answer does not by itself certify shortcut
retirement.}
\label{fig:case-p8}
\end{figure*}

\begin{figure*}[t]
\centering
\includegraphics[width=\textwidth]{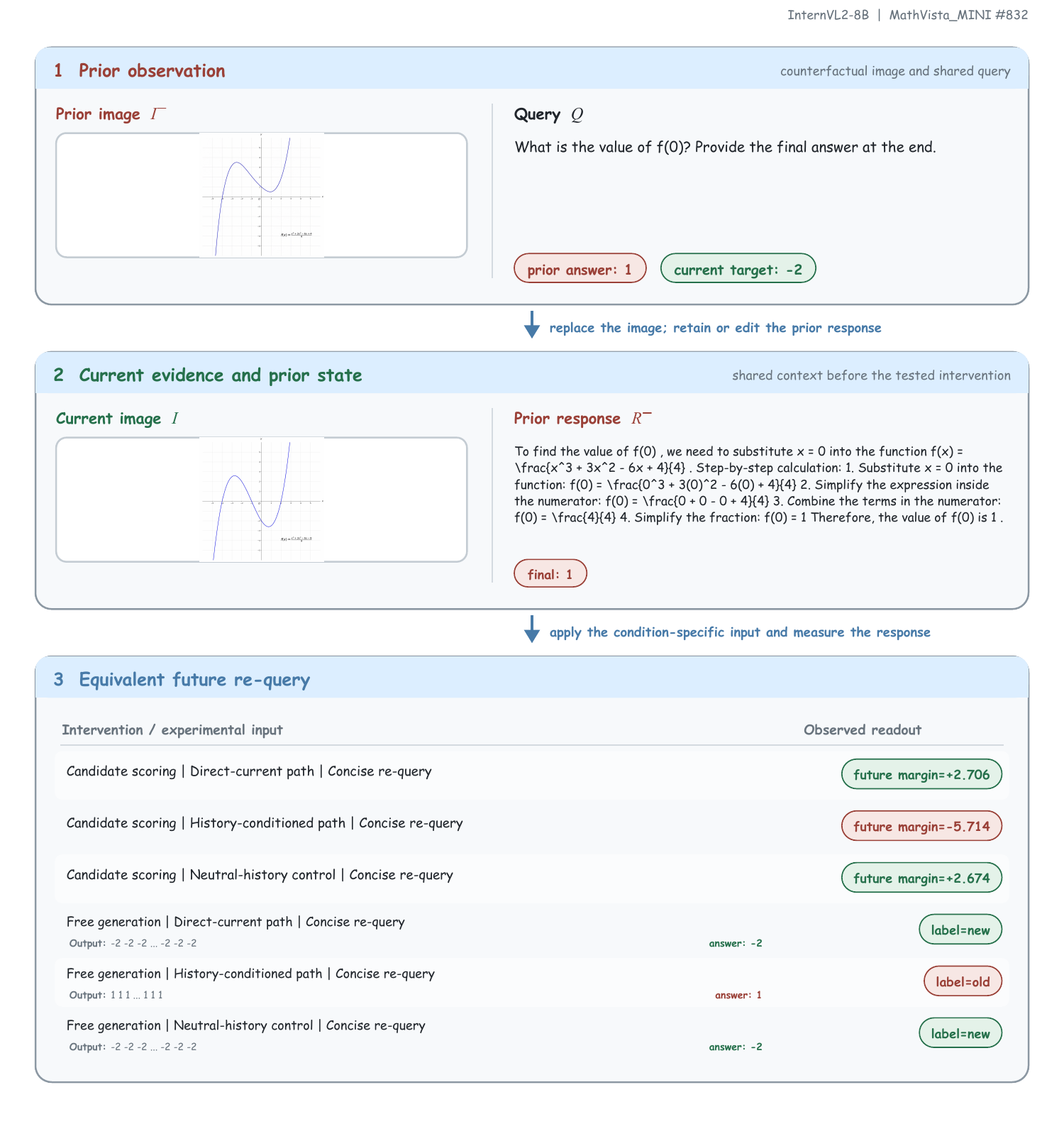}
\caption{\textbf{An active textual shortcut recurs under equivalent
re-query.}
After the image changes, InternVL2-8B receives the same concise future
question under three histories. The direct-current and neutral-history
paths favor the current answer, whereas the active stale-history path
favors the prior answer in both candidate scoring and free generation.
Holding the future query fixed illustrates the main-paper distinction:
answer recurrence is strong while the shortcut remains active, and is
not explained by a generic extra-turn effect.}
\label{fig:case-p10}
\end{figure*}

\begin{figure*}[t]
\centering
\includegraphics[width=\textwidth]{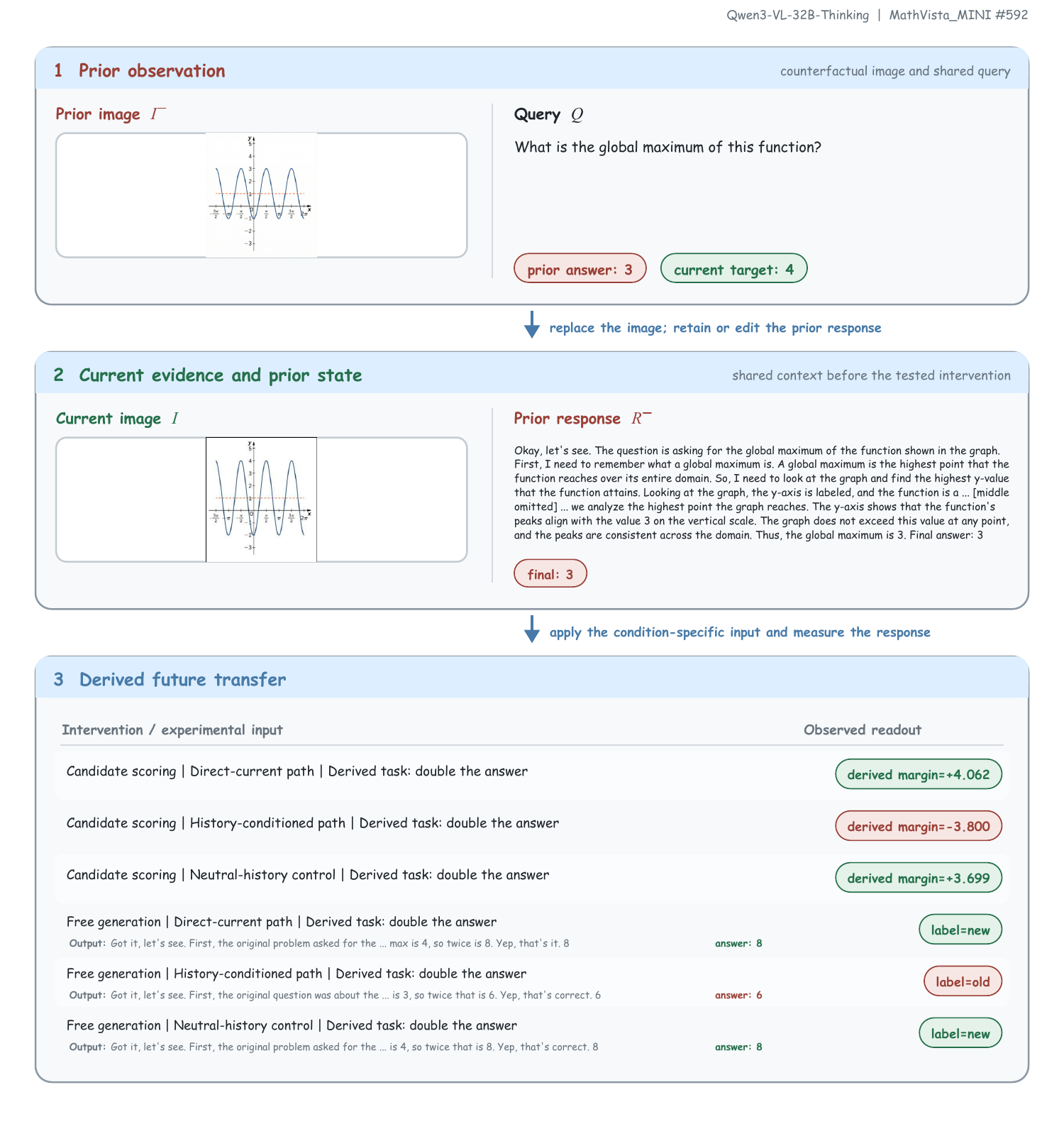}
\caption{\textbf{An active textual shortcut transfers a stale premise
into a new computation.}
Qwen3-VL-32B-Thinking is asked to double the answer after the image
change. The direct-current and neutral-history paths produce the
current-derived target, whereas the active stale-history path produces
the prior-derived target in both candidate scoring and free generation.
The case illustrates the main-paper finding that active stale influence
can propagate beyond answer repetition by supplying the premise for a
derived operation.}
\label{fig:case-p11}
\end{figure*}

The remaining cases ask whether prior-state influence has actually been
retired or can reappear in a later computation. They compare
direct-current, history-conditioned, and neutral-history paths under
matched support withdrawal, an identical future re-query, and a derived
task. Candidate margins and free-generation labels remain separate
readouts, allowing Figs.~\ref{fig:case-p8}--\ref{fig:case-p11} to expose
both shifts in answer preference and changes in the generated answer.

\end{document}